\documentclass{ecai} 

\usepackage{latexsym}
\usepackage{amssymb}
\usepackage{amsmath}
\usepackage{amsthm}
\usepackage{booktabs}
\usepackage{enumitem}
\usepackage{graphicx}
\usepackage{color}

\usepackage{makecell}

\theoremstyle{plain}

\theoremstyle{definition}

\theoremstyle{remark}

\usepackage{enumitem}
\usepackage[disable,textsize=tiny]{todonotes}

\usepackage{booktabs} 
\usepackage{algorithm}
\usepackage{algpseudocode}
\usepackage{subcaption}

\newcommand{\BibTeX}{B\kern-.05em{\sc i\kern-.025em b}\kern-.08em\TeX}

\begin{document}


\begin{frontmatter}


\paperid{4717} 


\title{One Shot vs. Iterative: Rethinking Pruning Strategies for Model Compression}


\author[A]{\fnms{Mikołaj}~\snm{Janusz}\thanks{Corresponding Author. Email: mikolaj1.janusz@student.uj.edu.pl}}
\author[A]{\fnms{Tomasz}~\snm{Wojnar}}
\author[B]{\fnms{Yawei}~\snm{Li}} 
\author[B]{\fnms{Luca}~\snm{Benini}}
\author[C]{\fnms{Kamil}~\snm{Adamczewski}}

\address[A]{Jagiellonian University} 
\address[B]{ETH Zürich}
\address[C]{Wrocław University of Science and Technology}


\begin{abstract}
Pruning is a core technique for compressing neural networks to improve computational efficiency. This process is typically approached in two ways: one-shot pruning, which involves a single pass of training and pruning, and iterative pruning, where pruning is performed over multiple cycles for potentially finer network refinement. Although iterative pruning has historically seen broader adoption, this preference is often assumed rather than rigorously tested. Our study presents one of the first systematic and comprehensive comparisons of these methods, providing rigorous definitions, benchmarking both across structured and unstructured settings, and applying different pruning criteria and modalities. We find that each method has specific advantages: one-shot pruning proves more effective at lower pruning ratios, while iterative pruning performs better at higher ratios. Building on these findings, we advocate for patience-based pruning and introduce a hybrid approach that can outperform traditional methods in certain scenarios, providing valuable insights for practitioners selecting a pruning strategy tailored to their goals and constraints. Source code is available at https://github.com/janumiko/pruning-benchmark.
\end{abstract}

\end{frontmatter}
\vspace{0.5cm}

\section{Introduction}

As the complexity and scale of tasks for machine learning and computer vision continue to grow, so does the size of neural networks designed to address these tasks~\cite{dosovitskiy2020image,he2015delving,he2016deep,huang2017densely, krizhevsky2012imagenet, simonyan2014very}. Larger models typically achieve superior performance. However, deploying these large models can be computationally expensive and resource-intensive, making them impractical for environments with limited computational power, such as mobile devices or embedded systems~\cite{han2015deep,han2016eie,lin2020mcunet}. Pruning is a critical technique that reduces network size without significantly compromising performance~\cite{chen2018constraint,ding2019centripetal, he2018amc,li2019oicsr,liu2019metapruning, liu2019rethinking,molchanov2019importance, ye2018rethinking}. By removing redundant parts of a network, pruning can help achieve results similar to those of fully trained networks, especially in inference scenarios where efficiency is essential.

\begin{figure}
    \centering
    \includegraphics[width=0.54\textwidth, trim=40 0 0 49, clip]{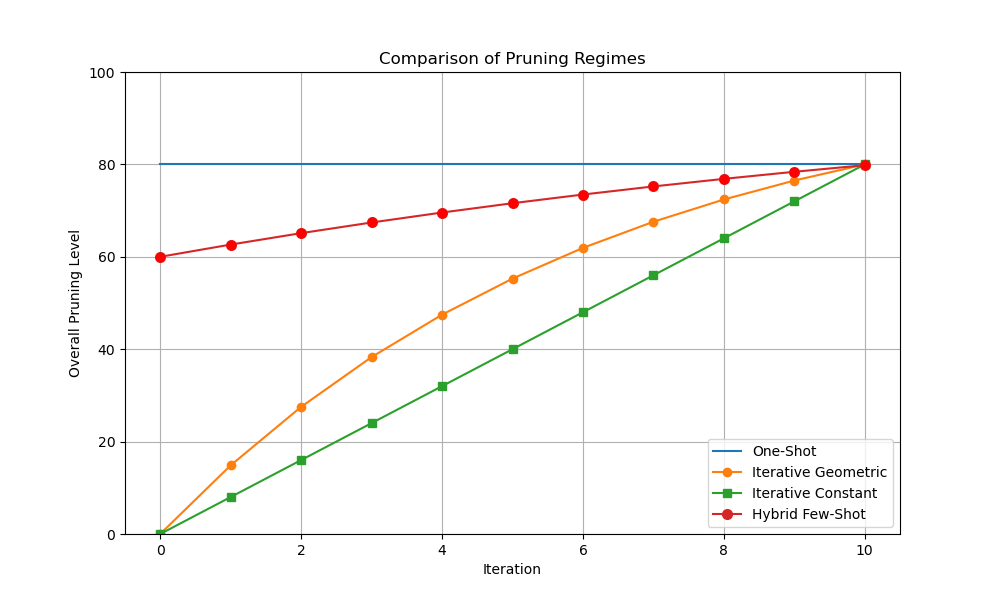}
    \caption{A conceptual illustration of the different pruning regimes. This schematic shows the progression of weight removal for one-shot, iterative constant, and iterative geometric pruning, and is not derived from a specific experiment.}
    \label{fig:pruning_regime}
    \vspace{0.8cm}
\end{figure}

Pruning techniques are broadly divided into two categories: \textbf{one-shot pruning} and \textbf{iterative pruning} (Figure~\ref{fig:pruning_regime}). One-shot pruning consists of a single cycle of training, pruning, and retraining, as detailed in Section~\ref{sec:pruning_regimes}. In contrast, iterative pruning involves multiple cycles of these steps, leading to potentially more refined and efficient network structures. Despite their prevalence, a systematic comparison of these two approaches under different regimes is limited in the literature. Prior research has largely focused on either the criteria for pruning~\cite{li2022revisiting, ye2018rethinking} or the specific pruning methods~\cite{cheng2023survey, liu2019rethinking,wang2021recent}, leaving a gap in understanding the relative performance of different pruning strategies under varying conditions.

In this work, we conduct a thorough comparison of one-shot versus iterative pruning across multiple settings. Such a comparison requires isolating the pruning regime as a variable, which fits into the broader view of pruning as a multi-objective optimization problem trading off accuracy, model size, and computational cost. Our controlled experimental setup holds the pruning criterion and target sparsity fixed to achieve this isolation. Furthermore, to properly evaluate the retraining cost, we propose patience-based fine-tuning as a more natural way to gauge fine-tuning duration. Our analysis focuses primarily on vision tasks, where the pruning literature is most extensive. To broaden our investigation, we also present an exploratory case study on a text generation task to explore how these regimes translate to the NLP domain.

For iterative pruning, we introduce a geometric pruning ratio scheduler in addition to the constant pruning ratio scheduler. Unlike the constant scheduler, which prunes a fixed percentage of weights across the entire network, the geometric scheduler prunes a fixed percentage of the remaining weights at each step, progressively removing fewer weights as pruning progresses. Our experiments reveal that the geometric pruning scheduler generally outperforms the constant scheduler.

Additionally, inspired by the respective strengths of one-shot and iterative pruning, we propose a \textbf{hybrid few-shot pruning} regime, combining aspects of both methods. This hybrid approach shows advantages over the individual techniques in certain scenarios. Notably, this study does not seek to establish one method as universally superior; instead, it aims to provide guidelines to help practitioners choose the most suitable pruning approach based on specific requirements and constraints.

To summarize, the contributions of this work include:
\begin{itemize}[leftmargin=*]
    \item Defining the problem of pruning regime selection, and the first broad comparison of one-shot and iterative pruning methods.
    \item The introduction of a geometric pruning ratio scheduler that prunes a fixed percentage of remaining weights in each pruning step, demonstrating benefits for iterative pruning.
    \item Proposing a hybrid approach that combines aspects of one-shot and iterative pruning to provide flexibility in specific cases.
    \item Broad empirical results that describe preferred settings for each pruning regime, in particular the impact of pruning regime on pruning criteria. 
\end{itemize}




\section{Pruning regimes in literature}
\label{sec:literature}

Pruning is one of the earliest and most established techniques for compressing neural networks. This section attempts to show the extent of the bias in the pruning literature towards proposing new pruning criteria. This bias underscores the importance of the proposed problem, which is broadly orthogonal and shows an alternative direction of pruning research.

Early research demonstrated that training large networks and then removing weights can significantly improve computational efficiency and reduce parameter size~\cite{hassibi1993second, lecun1990optimal}. Much of the pruning literature focuses on developing new methods to define pruning criteria, with the most common approaches based on the magnitude of weights~\cite{han2015deep, he2019filter}, sensitivity to derivatives~\cite{lecun1990optimal, molchanov2016pruning}, or empirical filter sensitivity~\cite{baykal2019sipping,liebenwein2019provable}. Other methods have been introduced that prune based on neuron similarity~\cite{NEURIPS2021_ce6babd0}, activation functions~\cite{lin2020hrank}, or even more unconventional criteria, such as game-theoretic~\cite{adamczewski2019neuron}, reinforcement learning~\cite{he2018amc}, or genetic algorithms~\cite{HANCOCK1992991,liu2019metapruning}.

Despite this diversity of pruning methods, there is no unified procedure for how networks are pruned. Instead, two main pruning strategies are prevalent in the literature: one-shot pruning and iterative pruning. These approaches differ in whether the network is pruned all at once or in successive stages. However, the impact of this choice on pruning outcomes is not well understood. Each approach also comes with a unique set of parameters that can significantly influence final network performance, such as fine-tuning duration and the number of pruning iterations. Details like the exact percentage of weights removed at each step are often not explicitly provided, leaving gaps in the understanding of how specific pruning configurations affect results. In this work, we argue that these choices are as crucial as pruning criteria and can greatly influence the overall performance of the pruned network. Our study aims to address this gap in the literature, conducting a detailed analysis to clarify how different pruning strategies affect network performance.

In the following sections, we discuss studies that employ both one-shot and iterative pruning, and in Table~\ref{tab:literature}, we summarize common pruning methods and their associated training protocols.


\paragraph{One-shot pruning.} One-shot pruning involves a single cycle of pruning followed by retraining, making it a computationally efficient approach widely appreciated in the literature. In this method, node ranking is computed only once, resulting in the final network structure after a single pruning step.

One-shot pruning is often chosen for methods where pruning costs are high. For instance, \citet{he2019filter} removes filters near the geometric median, arguing that these filters are effectively represented by the remaining ones. Calculating the geometric median, however, is computationally intensive. Similarly, Dirichlet pruning~\cite{adamczewski2020dirichlet} requires a training phase to compute parameters of the Dirichlet distribution, which serve as importance weights. CURL~\cite{luo2020neural}, on the other hand, removes all unimportant filters across layers in one step using a KL-divergence-based criterion. \citet{li2021heterogeneity} conducted one-shot pruning according to the gradients of the latent vectors in a hypernetwork, which is introduced to guide network pruning.

After the pruning step, retraining is typically performed. However, the optimal length for retraining has not been thoroughly investigated. Instead, many methods fix the retraining duration arbitrarily; for example, CURL retrains the model for 100 epochs.

\paragraph{Iterative pruning.} Iterative pruning refines neural network architectures by progressively removing less important parameters or structures while fine-tuning the network to maintain performance. Unlike one-shot pruning, which removes a significant portion of weights or neurons at once, iterative pruning involves multiple cycles of pruning and retraining.

Iterative pruning requires computing parameter importance at every step. Therefore, for large networks, iterative pruning can be computationally prohibitive in cases of unstructured pruning, due to the vast number of individual parameters. However, in smaller networks, iterative pruning dates back to early methods such as Optimal Brain Damage (OBD) and Optimal Brain Surgeon (OBS)~\cite{hassibi1993second}, which remove unimportant connections based on the second-order derivatives of the loss function with respect to the weights. 
In second-derivative methods, \cite{hassibi1993second} requires calculating the inverse Hessian matrix, which involves significant computational effort. Given the computational cost of calculating such pruning criteria, only one~\cite{theis2018faster} or a few batches~\cite{liebenwein2019provable} are often used to evaluate parameter sensitivity.

In iterative pruning, a common approach is to use a constant pruning rate—a fixed percentage of weights pruned in each cycle. However, this rate varies across methods. For example, \cite{molchanov2019importance} uses Taylor expansions to approximate filter contributions, removing 10 neurons every 30 mini-batches until the target number of pruned neurons is reached, followed by fine-tuning for 25 epochs. In~\cite{renda2020comparing}, 20\% of the lowest-magnitude weights are pruned globally, after which the network is retrained using learning rate rewinding and the original training time. \cite{li2022revisiting} explores sub-architecture optimization by randomly removing 10\% of weights, followed by fine-tuning.

The number of elements pruned in each cycle may be determined by various criteria, often treated as hyperparameters. \cite{han2015learning} sets a custom threshold, removing elements whose magnitude is below that threshold. \cite{liu2021group} prunes channels at each iteration based on Fisher Information scores, allowing the number of pruned channels to vary. \cite{liebenwein2019provable} uses a hyperharmonic sequence for pruning ratios, where the i-th pruning ratio follows \( 1 - \frac{1}{(i+1)^\alpha} \). \cite{lin2020hrank} prunes a specified number of channels layer-wise, fine-tuning after each layer’s pruning. Additionally, \cite{luo2017thinet} uses statistics from subsequent layers to prune each layer iteratively, fine-tuning for one or two epochs after each layer is pruned. ~\citet{li2020group} applied group sparsity to the sparsity-inducing matrix and conducted a proximal gradient descent algorithm to progressively prune the network during the optimization of the pruning procedure.

\paragraph{Pruning at initialization and during training.} 
While this work primarily focuses on post-training pruning using a train-prune-retrain cycle, we acknowledge other types of pruning. Pruning at initialization~\cite{frankle2018lottery, lee2018snip, Synflow} avoids pruning a fully trained model, instead identifying a smaller subnetwork to train from scratch for comparable performance to the larger model.

Another category involves pruning during training through regularization, which encourages certain parameters to approach zero~\cite{oh2020radial, wang2021convolutional}. For example, \cite{wang2021convolutional} introduces scaling factors that selectively scale the outputs of certain CNN structures, applying sparse regularization to these scaling factors to progressively reduce their influence during training.





\begin{table*}[]
    \centering
\small
\begin{tabular}{cccc}
    \toprule
    Method & structure & regime & step \\ \midrule
    HRank \cite{lin2020hrank} & structured & iterative (custom) & \\
    ThiNet \cite{luo2017thinet} & structured & iterative (custom) & \\
    SSS \cite{huang2018data} & structured & iterative (unspecified) & \\
    Revisiting Random Pruning (RRCP) \cite{li2022revisiting} & structured & iterative (constant) & 10\%\\

    Fisher information \cite{theis2018faster} & structured & iterative & \\ 
    Learning rate rewinding \cite{renda2020comparing} & both & iterative (constant) & 20\%\\
    Optimal brain damage \cite{lecun1990optimal} & unstructured & iterative & \\
    Optimal brain surgeon \cite{hassibi1993second} & unstructured & iterative & \\
    Learning weights and connections \cite{han2015learning} & unstructured & iterative & \\
    Taylor Expansion \cite{molchanov2019importance} & structured & iterative (constant) & 2\% \\
    Group Fisher Information \cite{li2020group} & structured & iterative (custom) & \\
    Empirical Sensitivity Analysis \cite{liebenwein2019provable} & structured & iterative (custom) & \\  
    CURL (KL-divergence metric) \cite{luo2020neural} & structured& one-shot\\ 
    Dirichlet Pruning \cite{adamczewski2020dirichlet} & structured& one-shot &\\
    Geometric Median \cite{he2019filter} & structured & one-shot & \\
    \bottomrule

    \end{tabular}
          \vspace{0.5cm}
    \caption{A list of selected pruning methods and their corresponding training regimes}
\vspace{0.7cm}
    \label{tab:literature}
\end{table*}

\paragraph{Unstructured and structured pruning.} Pruning networks can be done for individual weights or for structures within the network. Each of them is important in its own regard and is broadly researched. Hence, this benchmark includes tests where both unstructured and structured pruning are considered. We build our benchmark around the Torch-Pruning \cite{fang2023depgraph} for structured pruning, and extend it with unstructured pruning using the built-in pruning utilities in PyTorch. The details for how unstructured and structured pruning are done can be found in the Appendix.

\section{Pruning regimes}
\label{sec:pruning_regimes}

In the context of neural network optimization, pruning regimes can be split into categories based on the following question: \textit{How should we divide the pruning process?} Should we prune all the redundant weights at once? Or should we divide the pruning between different iterations that are separated by modifying the structure of the network and weight updates? In this section, we present rigorous definitions of pruning criteria.

For the purpose of the comparison of the pruning regimes, assume that \( W \) is the total number of weights in the neural network and \( p \) is the desired pruning percentage, e.g. \( p = 0.8\) means a pruning of 80\% of weights.
\begin{itemize}[leftmargin=*]
    \item \textbf{One-shot Pruning:} One-shot pruning is a regime where a substantial portion of the network's weights are removed in a single pruning step, following the process:
    \begin{itemize}
        \item  Initial Training: The neural network is first fully trained to ensure it learns the patterns in the data.
        \item Pruning Step: A certain percentage of the least important weights or neurons (based on metrics like magnitude or importance) are pruned in one go.
        \item Fine-tuning: After pruning, the network may undergo additional fine-tuning to regain any performance lost from pruning.
    \end{itemize}
The least significant weights are eliminated based on a predetermined importance criterion. The number of weights removed after one-shot pruning is given by:
        \[ p \times W. \] 

    \item \textbf{Iterative Pruning:} Iterative pruning is a process where weights are pruned over multiple iterations, allowing the network to gradually retrain and recover some performance loss. In each iteration the ranking, pruning, and fine-tuning occurs. We define two common approaches to iterative pruning and name them differently to avoid confusion.
    
    \begin{itemize}
    
        \item \textbf{Iterative Constant:} A constant number of parameters is pruned at each step. Let \( \textit{steps} \) be the number of iterations, then $$ \frac{p \times W}{\text{steps}} $$ weights are pruned at each stage. The pruned percentage per step is fixed in relation to the initial number of weights.
        
        \item \textbf{Iterative Geometric}:
    A fixed percentage \( p \) of the remaining weights is pruned at each step, meaning that as the pruning process progresses, fewer weights are pruned at each iteration.
    
    The number of weights at step $n$ is given by the following formula:
    $$ W \times (1 - \frac{p}{\text{steps}})^{n} $$
    \end{itemize}


\item \textbf{Hybrid pruning:} We propose a new pruning regime, hybrid (few-shot) pruning, which is a combination of the idea of one-shot and geometric regimes. The majority of the weights are removed at the first step and the model is retrained for a longer time. Then for the remaining weights, we perform a more fine-grained geometric-like pruning over several iterations. The hybrid pruning approach can be summarized as follows:
\begin{itemize}
    \item Apply large pruning ratio $p_k$ to the original network and perform a longer fine-tuning phase.
    \item Perform iterative geometric pruning with smaller pruning ratio, $p_i \ll p_k$ starting from the state with $p_k$ parameters removed. Proceed until the desired final pruning percentage $p$.
\end{itemize}
\end{itemize}

We elaborate on the hybrid regime and empirical estimates for $p_k$ and $p_i$ in Section~\ref{sec:hybrid}. 

\section{Pruning regime factors}

Pruning regimes, whether one-shot or iterative, involve parameters that significantly impact their effectiveness. To properly evaluate these regimes, it is essential to consider these parameters.

\subsection{Retraining}

Retraining, or fine-tuning, is a critical step in the pruning process, helping the neural network recover performance after a portion of its weights or channels has been removed. Retraining is necessary because the initial pruning can degrade the model’s accuracy by eliminating parameters that the network previously relied upon for making predictions.

Both one-shot and iterative pruning require retraining, though the duration may vary. Generally, the larger the drop in accuracy, the longer the retraining phase should be. The accuracy drop typically correlates with the number and importance of pruned parameters, leading to longer fine-tuning for one-shot pruning and shorter, repeated fine-tuning phases for each step in iterative pruning.

However, the optimal length of retraining has not been rigorously studied, with methods often using arbitrary values. For instance, \cite{renda2020comparing} fine-tunes the model for the full original training time, which may be computationally excessive since the network is not being trained from scratch. Conversely, in \cite{crowley2018closer, luo2017thinet}, fine-tuning is limited to one epoch after each pruning step in iterative pruning.

\paragraph{Patience (early stopping).} In this work, we advocate for using patience or early stopping to determine the number of fine-tuning epochs in both one-shot and iterative pruning. Patience allows a dynamic approach to fine-tuning, where the model is trained until a specified criterion (e.g., validation accuracy) no longer improves over a set number of epochs. The best-performing checkpoint is retained. Patience is beneficial because it adapts the fine-tuning duration based on actual performance, whereas fixed epochs may either be insufficient or wasteful. In this study, we use patience as an alternative to a fixed epoch count, defining it as the number of epochs to continue fine-tuning before stopping if no improvement occurs. The details of our patience-based algorithm are outlined in Appendix Algorithm 1. In Appendix Section~\ref{sec:ablation} we present a comparative study to emphasize the importance of proper selection of both patience-based retraining length and step size.

\subsection{Iteration Pruning Rate}

In iterative pruning, a key parameter is the iteration pruning rate, which determines the percentage of parameters removed at each step. Assuming a fixed target pruning rate, the iteration pruning rate is defined either by the number of steps or, conversely, the number of steps is determined based on the selected pruning rate.

As discussed in Section~\ref{tab:literature}, the pruning rate varies widely across methods, from as low as 1\% to as high as 20–30\%. Reflecting on these values, we examine both single-digit and double-digit iterative pruning rates for constant and geometric pruning. In constant pruning, the iterative rate should evenly divide the final pruning rate. In geometric pruning, however, the amount pruned at each step decreases, and the rate is set so that the geometric sum of pruning rates across steps achieves a predefined final pruning level. Further details can be found in the Appendix.

\begin{figure*}[tp]
\centering
\begin{subfigure}[b]{0.32\textwidth}
    \centering
    \includegraphics[width=\textwidth, trim=15 3 80 65, clip]{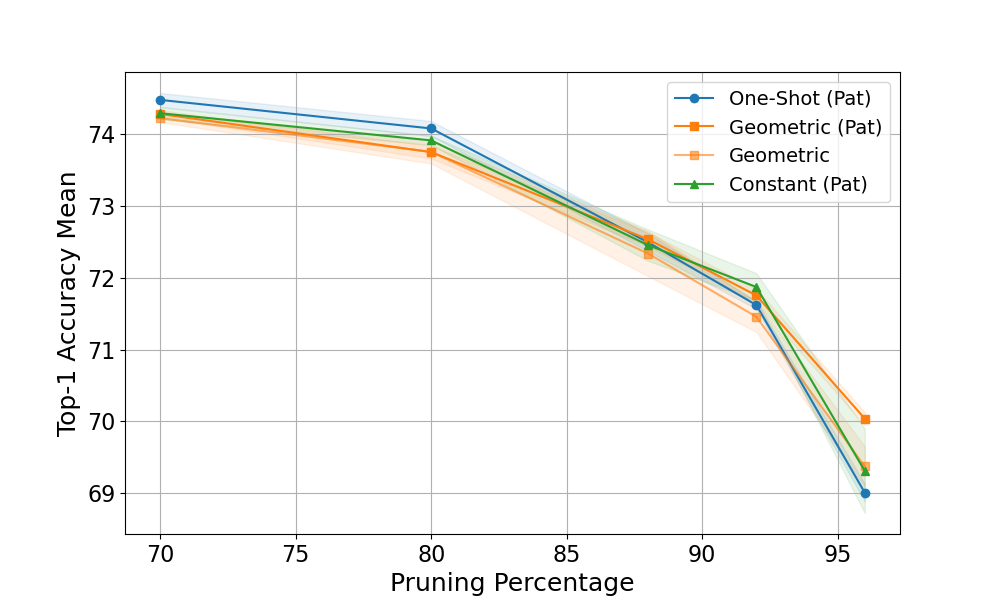}
    \caption{\makecell{ResNet-18 / CIFAR-100 \\ }}
\end{subfigure}
\hfill
\begin{subfigure}[b]{0.32\textwidth}
    \centering
    \includegraphics[width=\textwidth, trim=15 3 80 65, clip]{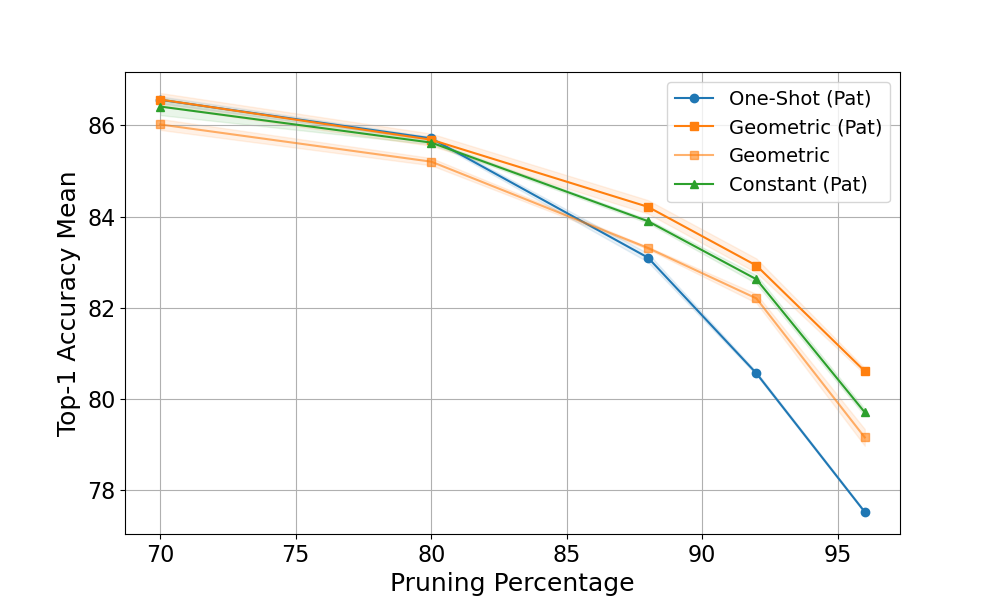}
    \caption{\makecell{EfficientNet / CIFAR-100 \\ }}
\end{subfigure}
\hfill
\begin{subfigure}[b]{0.32\textwidth}
    \centering
    \includegraphics[width=\textwidth, trim=15 3 80 65, clip]{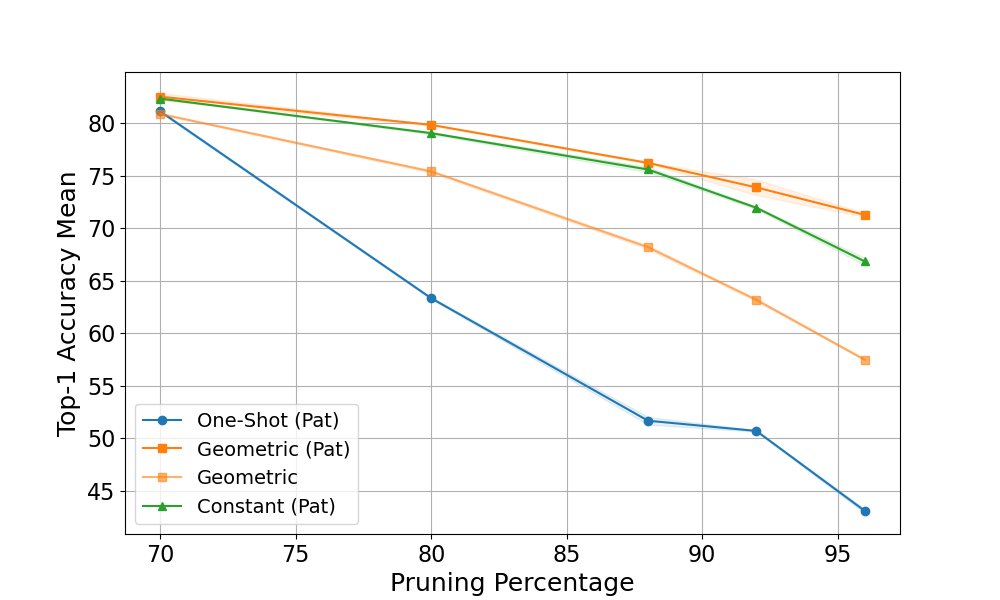}
    \caption{\makecell{ViT / CIFAR-100 \\ }}
\end{subfigure}
\vspace{1.3cm}

\begin{subfigure}[b]{0.32\textwidth}
    \centering
    \includegraphics[width=\textwidth, trim=15 3 80 65, clip]{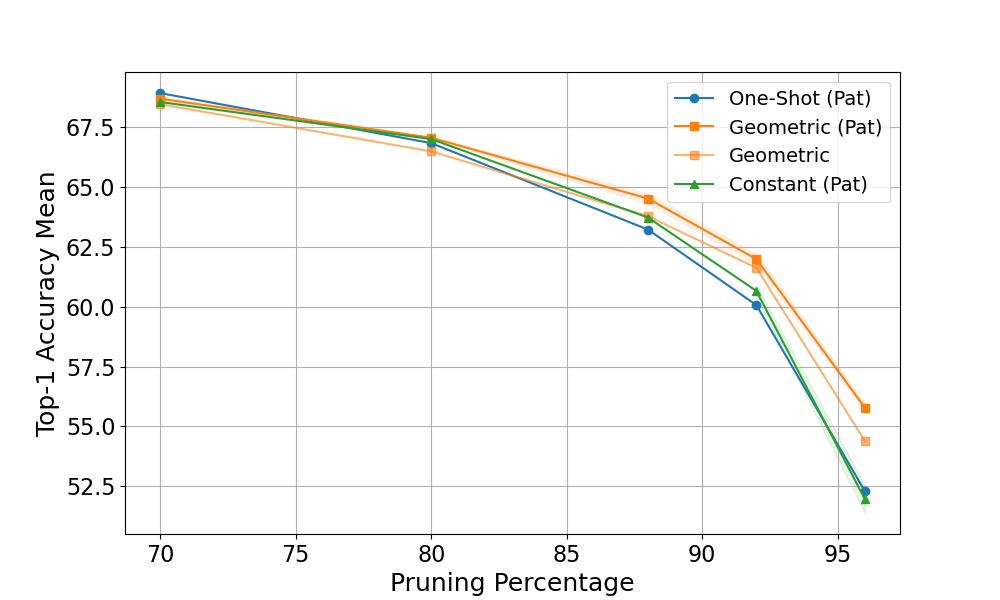}
    \caption{\makecell{ResNet-18 / Imagenet \\ }}
\end{subfigure}
\hfill
\begin{subfigure}[b]{0.32\textwidth}
    \centering
    \includegraphics[width=\textwidth, trim=15 3 80 65, clip]{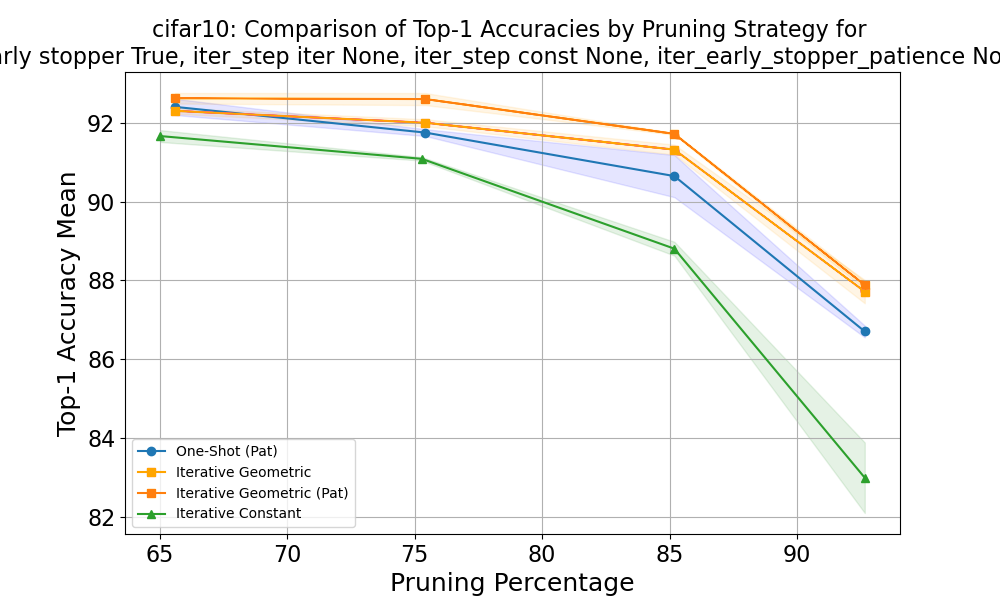}
    \caption{\makecell{ResNet-18 / CIFAR-10 \\ (Structured)}}
\end{subfigure}
\hfill
\begin{subfigure}[b]{0.32\textwidth}
    \centering
    \includegraphics[width=\textwidth, trim=15 3 80 60, clip]{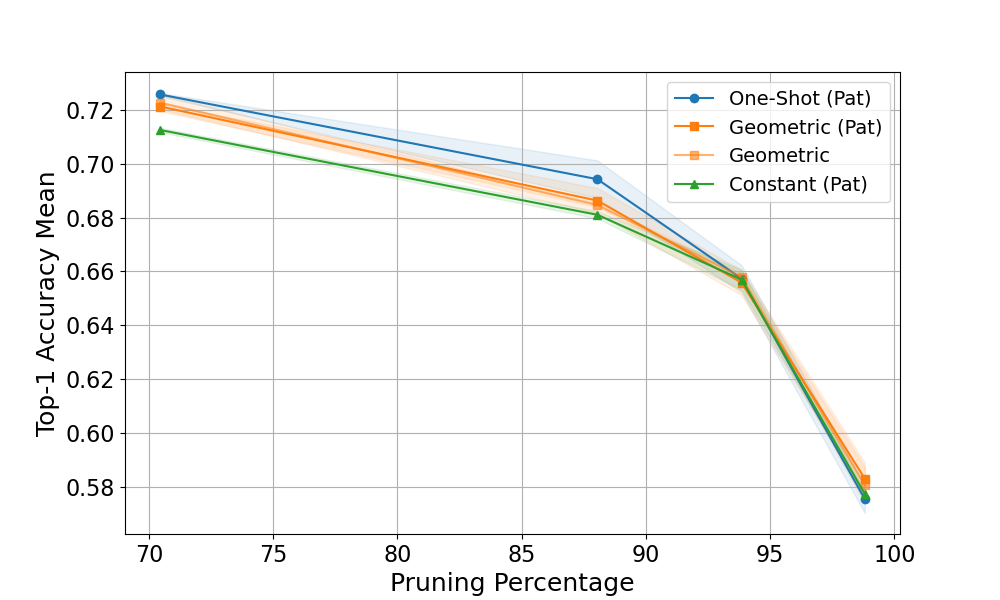}
    \caption{\makecell{ResNet-18 / CIFAR-100 \\ (Structured)}}
\end{subfigure}
\vspace{0.8cm}
\caption{Comparison of pruning regimes across architectures and datasets. Method with (Pat) in the name indicate the patience-based fine-tuning. The performance of one-shot, iterative constant and iterative geometric regimes are plotted. 'Geometric' outperforms 'Constant' in most high-sparsity scenarios. The y-axis represents Top-1 Accuracy (\%). See Appendix for fixed-length regimes.}
\label{fig:main_results}
\vspace{0.8cm}
\end{figure*}

\section{Pruning regime empirical evaluation}
\label{sec:empirical}

This section consists of five parts. We first present some main takeaways and then present the detailed experiments. In the second part, we present broad results in Computer Vision and Natural Language Processing settings for the most common magnitude pruning. In the third part, we extend the results to other pruning criteria and  highlight that the choice of pruning regime impacts the pruning outcomes differently when different pruning criteria are applied, a problem broadly overlooked in the literature. In the fourth section, we consider comparing the regimes when the pruning computational budget is fixed. In the final part, based on our analysis of one-shot and iterative regimes, we propose a novel hybrid approach that combines both existing pruning regimes, retaining its strength and producing a more informed and better-performing pruning regime.

\paragraph{Experimental set-up.} We perform experiments on several datasets and model architectures. The datasets include vision datasets, CIFAR-10 \cite{cifar10}, CIFAR-100, and Imagenet1K \cite{deng2009imagenet} and the language dataset TinyStories \cite{eldan2023tinystories}. The experiments are performed both on convolutional neural networks and transformers, in particular ResNet~\cite{he2016deep}, EfficientNet~\cite{tan2019efficientnet}, Visual Transformer~\cite{dosovitskiy2020image} and TinyStories-33M ~\cite{eldan2023tinystories}. The open-sourced codebase allows for other custom choices.  As recommended in \cite{han2015learning}, we use $1/10^{\text{th}}$ of the original learning rate for the fine-tuning phase.

\subsection{Key observations}

\begin{itemize}
\setlength{\itemsep}{2pt}  
    \setlength{\parskip}{2pt}  
    \item One-shot pruning can perform better than iterative pruning for CNNs and lower pruning rates, and iterative pruning is better for transformers and higher rates.
    \item One-shot pruning typically reduces retraining time compared to iterative pruning by avoiding repeated cycles of pruning and retraining, which is especially helpful when computational resources are limited.
    \item Iterative geometric pruning is superior to constant iterative pruning in most cases.
    \item Early stopping ensures optimal fine-tuning time.
    \item Number of retraining iterations matters significantly.
    \item Iterative pruning is preferable for second-derivative methods.
\end{itemize}

\subsection{Comparison of one-shot and iterative regimes.}

\paragraph{One-shot pruning with patience-based retraining.} We demonstrate that one-shot pruning, when paired with an adaptive retraining duration, can be highly effective, surpassing both forms of iterative pruning, as shown in Fig.~\ref{fig:main_results}. Our approach to one-shot pruning uses patience-based retraining, allowing the model to stop fine-tuning once there is no improvement over a specified number of epochs. This method is more adaptive than using a fixed number of epochs, which may result in either insufficient or excessive retraining. Notably, one-shot pruning consistently outperforms iterative pruning, particularly at pruning rates below 80\%.

\paragraph{Iterative pruning with fixed retraining.} In contrast, iterative pruning in the literature is often paired with a fixed fine-tuning phase, sometimes limited to as little as one epoch~\cite{crowley2018closer}. In our experiments, we test a range of fixed retraining durations for both iterative geometric and iterative constant pruning, selecting the best-performing configuration, which is plotted in Figure~\ref{fig:main_results}. Results indicate that short, fixed retraining phases in iterative pruning lead to suboptimal performance. Geometric iterative pruning performs better at higher pruning ratios and in transformer models and structured pruning contexts.

\paragraph{Iterative pruning with patience-based retraining.} To enable a fair comparison with one-shot pruning, we propose implementing patience-based fine-tuning for iterative pruning, allowing both methods to benefit from early stopping. Since iterative pruning removes smaller fractions of weights in each step, we set a shorter patience period than in one-shot pruning to maintain efficiency. An ablation study on patience values is included in Section~\ref{sec:ablation}. As shown in Figure~\ref{fig:main_results}, patience-based iterative pruning improves fine-tuning effectiveness over standard iterative pruning, achieving higher accuracies, especially at high pruning ratios where heavily pruned networks are more sensitive to overtraining or undertraining. Iterative pruning is also preferable for transformers.

\begin{figure*}[h]
\centering

\begin{subfigure}[b]{0.32\textwidth}
    \centering
    \includegraphics[width=\textwidth, trim=0 0 40 50, clip]{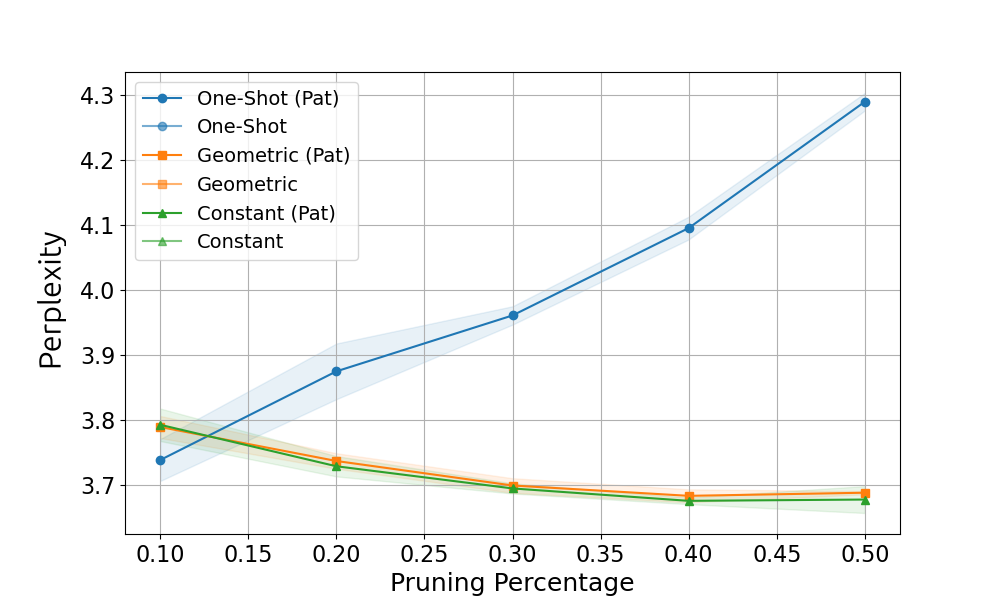}
    \caption{TinyStories/TinyStories-33M (GPT-Neo)}
\end{subfigure}
\hfill
\begin{subfigure}[b]{0.32\textwidth}
    \centering
    \includegraphics[width=\textwidth, trim=0 0 40 35, clip]{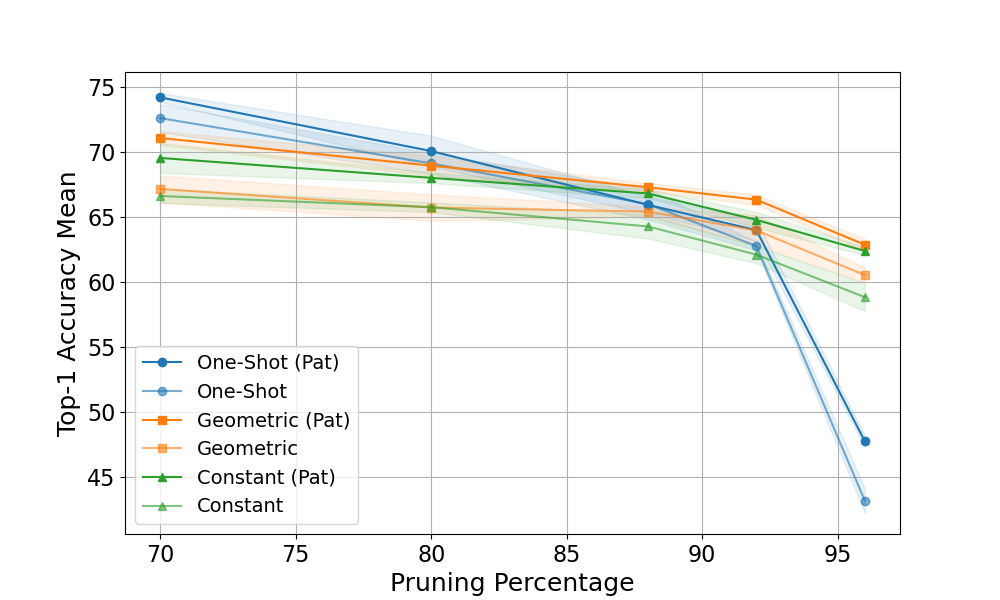}
    \caption{ResNet-18 / CIFAR-100 / Hessian }
\end{subfigure}
\hfill
\begin{subfigure}[b]{0.32\textwidth}
    \centering
    \includegraphics[width=\textwidth, trim=0 0 40 35, clip]{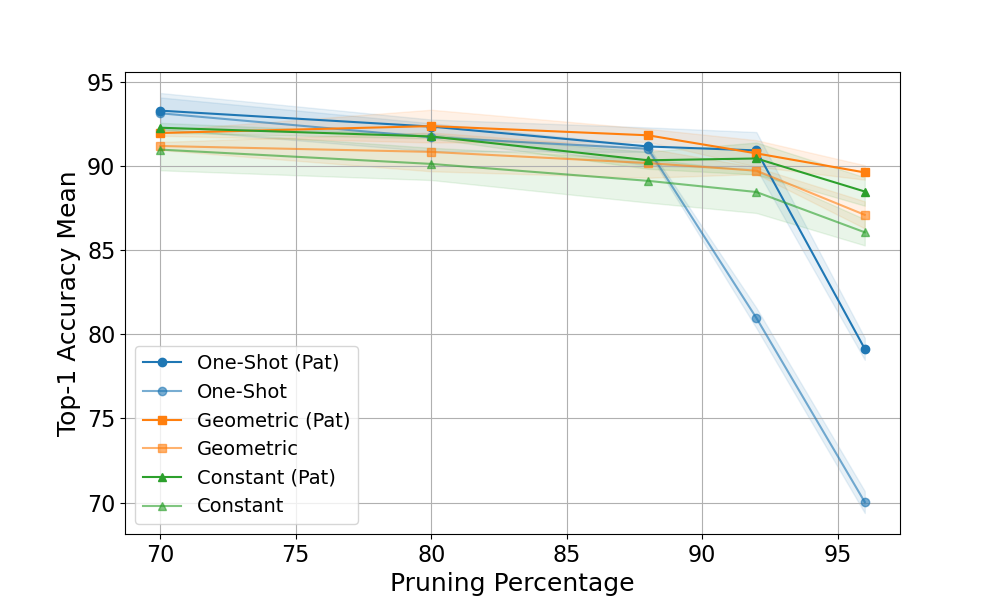}
    \caption{ResNet-18 / CIFAR-10 / Hessian}
\end{subfigure}
\vspace{0.8cm}
\caption{The performance of training regimes for (a) natural language processing TinyStories text generation dataset. Lower perplexity means better performance. (b-c) second-derivative pruning criteria on vision datasets.}
\label{fig:tiny_crit}
\vspace{0.8cm}
\end{figure*}

\begin{figure*}[h]
\centering

\begin{subfigure}[b]{0.32\textwidth}
    \centering
    \includegraphics[width=\textwidth, trim=0 0 70 65, clip]{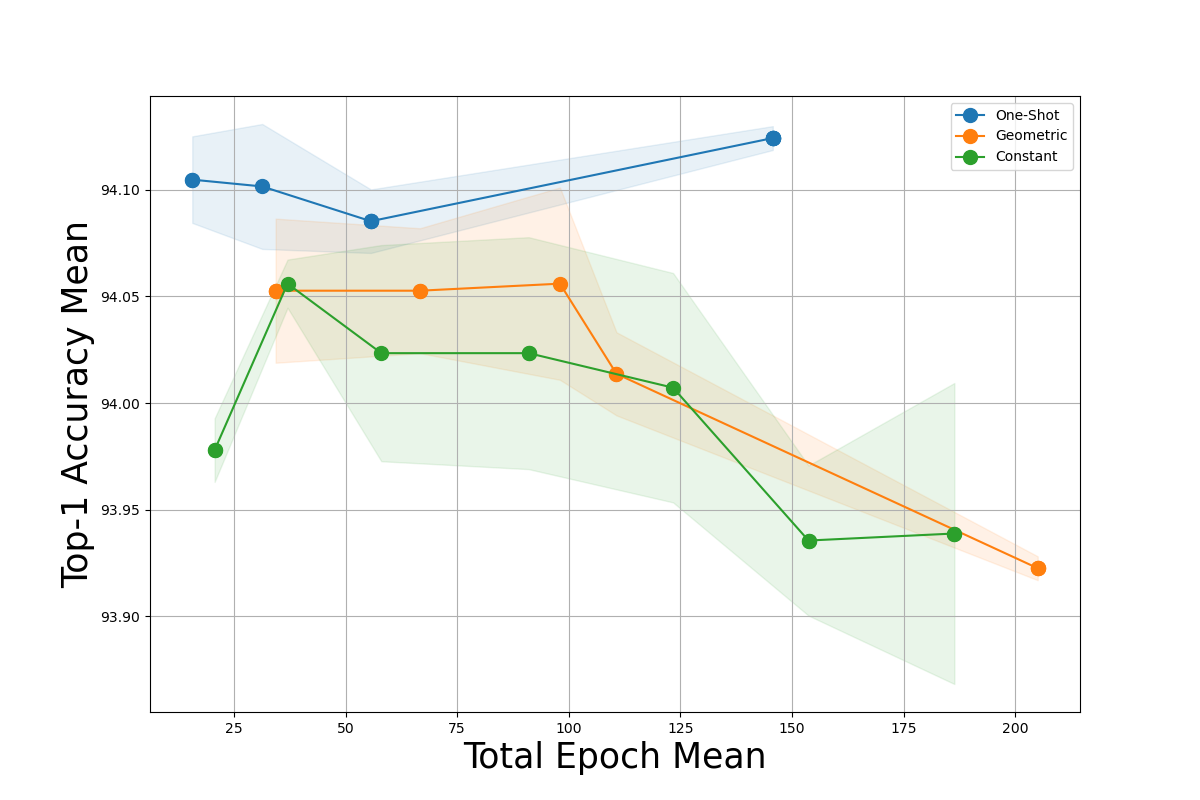}
    \caption{Pruning rate 70\%}
\end{subfigure}
\hfill
\begin{subfigure}[b]{0.32\textwidth}
    \centering
    \includegraphics[width=\textwidth, trim=0 0 70 65, clip]{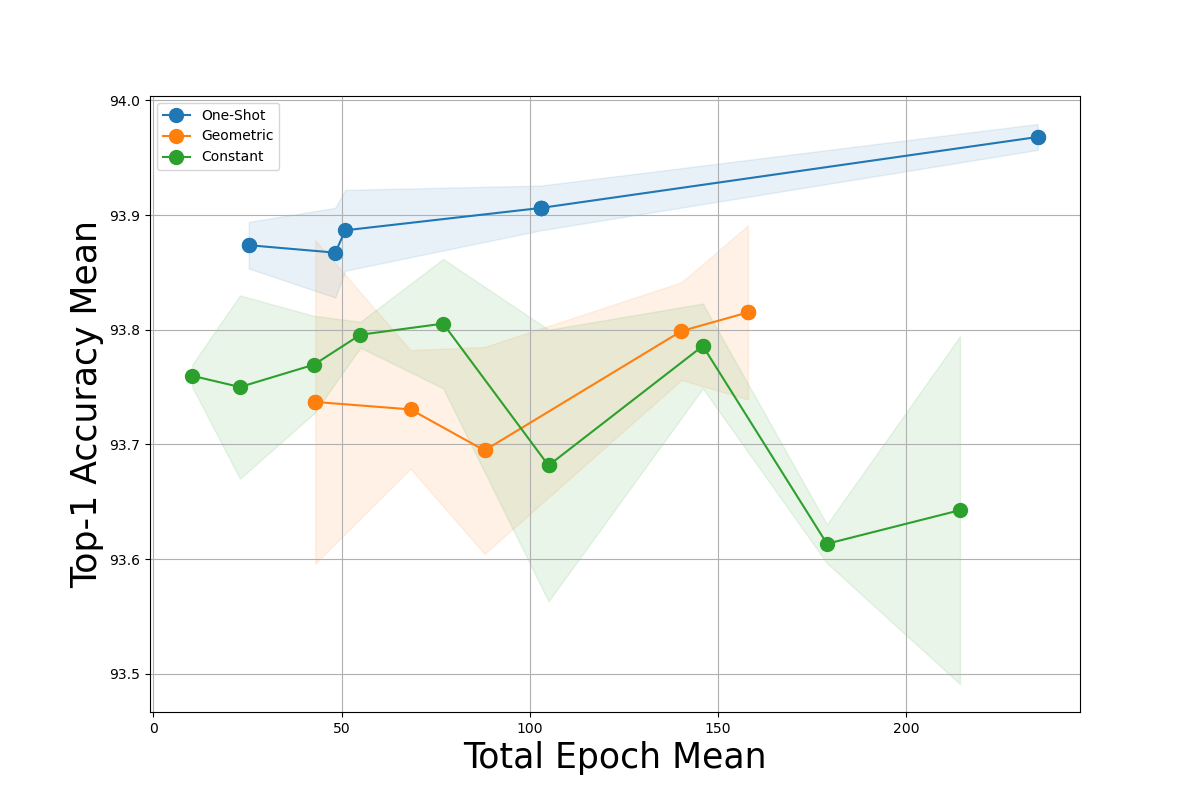}
    \caption{Pruning rate 80\%}
\end{subfigure}
\hfill
\begin{subfigure}[b]{0.32\textwidth}
    \centering
    \includegraphics[width=\textwidth, trim=0 0 70 65, clip]{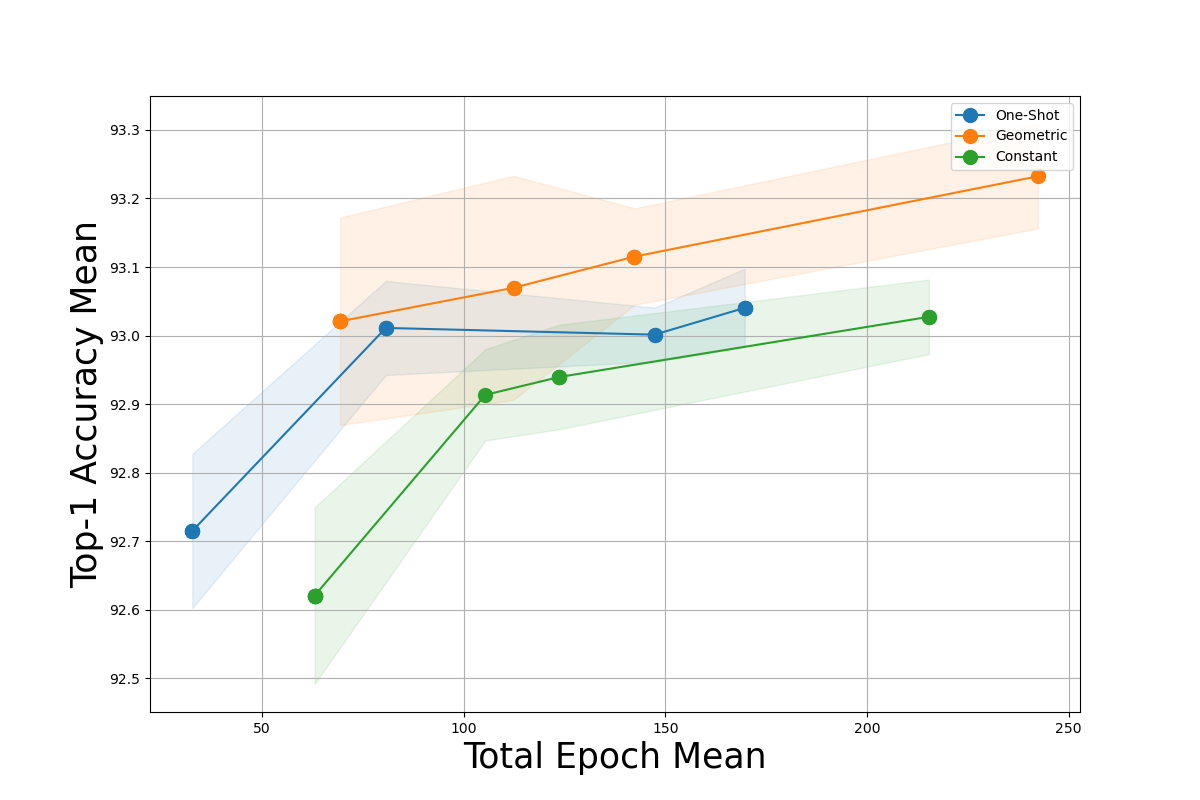}
    \caption{Pruning rate 92\%}
\end{subfigure}
\vspace{0.8cm}

\caption{The performance of training regimes for fixed computational budget, given in terms of total number of epochs. One-shot is more efficient for pruning rates below 80\% while iterative geometric for higher pruning rates.}
\label{fig:retraining_budget}
\vspace{0.8cm}
\end{figure*}

\begin{figure}[th]
    \centering
    \includegraphics[trim=40 0 0 0, clip, width=1.1\linewidth, height=5cm]{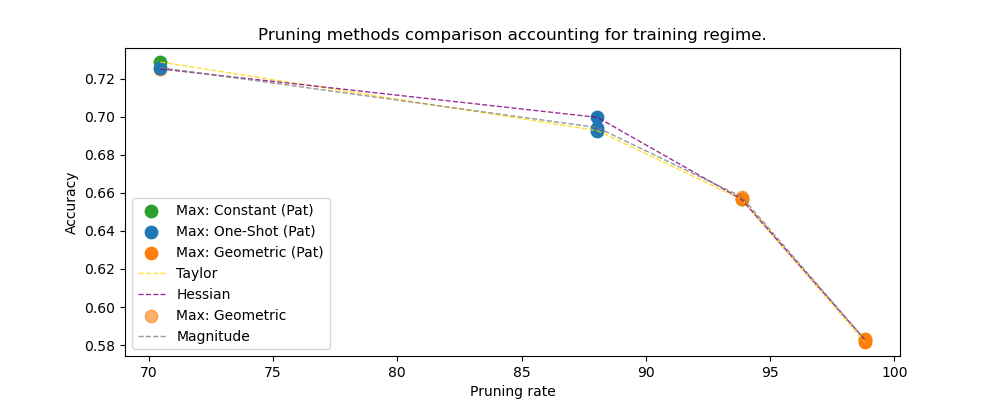}
    \vspace{0.3cm}
    \caption{Each dashed line tracks the performance of a single pruning criterion. The coloured dot on the line indicates which regime (one-shot, iterative, etc.) achieved that best result at a given pruning ratio.}
 \vspace{0.8cm}
    \label{fig:criteria_comparison}
    \vspace{0.8cm}
\end{figure}

\subsubsection{Natural language processing}

In addition to computer vision tasks, we also conduct experiments on a natural language processing (NLP) task, specifically text generation. For these experiments, we prune the pre-trained TinyStories-33M language model~\cite{eldan2023tinystoriessmalllanguagemodels}, which is based on GPT-Neo~\cite{gpt-neo}. We use the perplexity metric to evaluate pruning and fine-tuning on the TinyStories dataset. Perplexity measures how well a probabilistic model predicts a sequence of words, with lower perplexity indicating better performance. As in our previous experiments, we explore various pruning schedules and apply patience-based fine-tuning. The results are shown in Fig.~\ref{fig:tiny_crit}. Generally, we observe a similar relative performance pattern between pruning regimes: one-shot pruning performs better at lower pruning ratios, while iterative pruning excels at higher compression rates, with iterative pruning showing a notably larger advantage in this context. However, unlike vision tasks, in case NLP models are more sensitive to one-shot pruning, showing performance degradation even when only 10–20\% of the parameters are removed. On the other hand, interestingly, we find that in the case of iterative pruning, perplexity decreases as pruning progresses, suggesting that the LLM contains a substantial number of redundant parameters and benefits from pruning.

\subsection{Pruning methods comparison}

\begin{figure*}[th]
\centering

\begin{subfigure}[b]{0.46\textwidth}
    \centering
    \includegraphics[width=\textwidth, trim=0 0 70 50, clip]{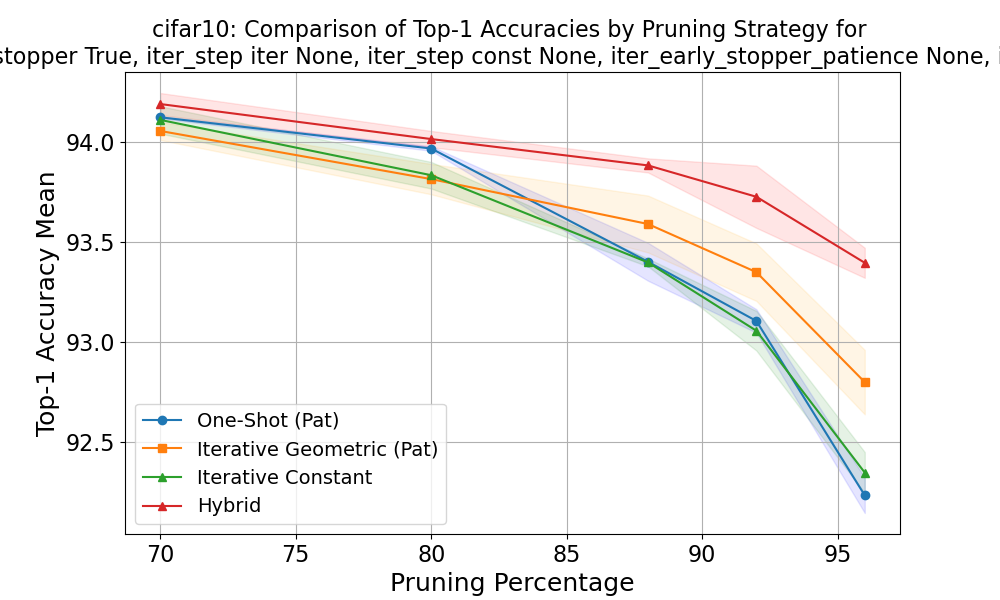}
    \caption{ResNet-18 / CIFAR-10}
\end{subfigure}
\hfill
\begin{subfigure}[b]{0.46\textwidth}
    \centering
    \includegraphics[width=\textwidth, trim=0 0 70 50, clip]{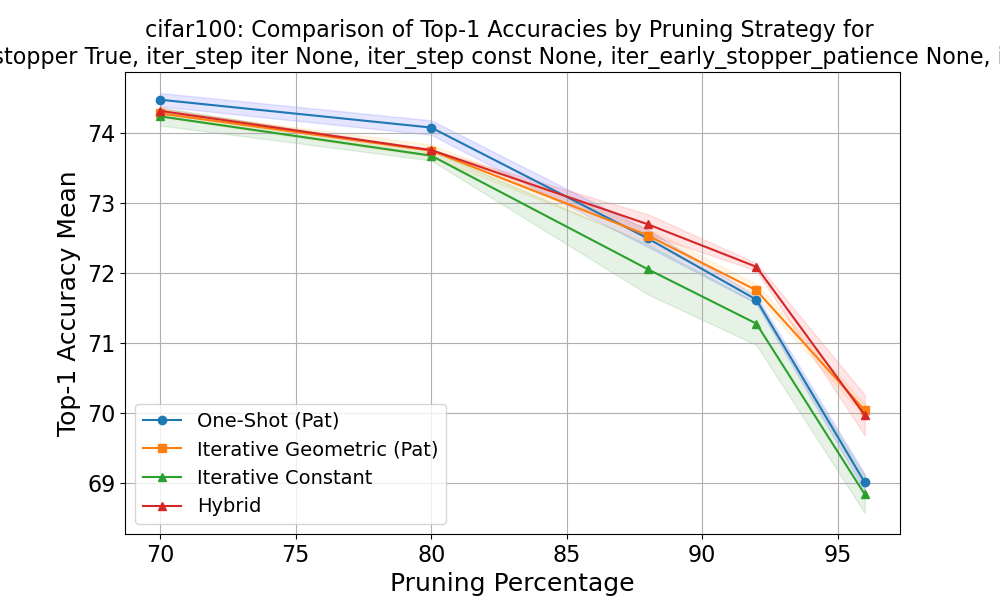}
    \caption{ResNet-18 / CIFAR-100}
\end{subfigure}

\vspace{0.8cm}
\caption{Hybrid approach in comparison with one-shot and iterative pruning.}
\label{fig:hybrid}
\vspace{0.8cm}
\end{figure*}


We then examine how the choice of pruning criteria influences the selection of a training regime. In this study, we compare three key criteria: magnitude-based, Taylor Expansion, and Hessian-based pruning. The results are presented in Fig.~\ref{fig:criteria_comparison}.

Generally, for lower pruning ratios, the one-shot regime performs better across all criteria except the constant regime. Notably, second-order approaches outperform Taylor Expansion at 70\% pruning and Hessian-based pruning at 88\%. However, at higher pruning rates (over 90\%), an interesting conclusion emerges: the pruning regime becomes less significant, as all criteria yield similar performance. In these cases, the iterative geometric approach is preferred across the board.

From a computational perspective, these findings are encouraging. The cost of computing pruning rankings varies: it is lowest for magnitude-based pruning and highest for second-order approaches. Since second-order pruning performs better at lower pruning ratios, it is computationally efficient in one-shot scenarios, as the rankings need to be computed only once. Conversely, for higher pruning ratios where iterative pruning is preferable, the choice of criterion becomes less critical. In such cases, magnitude-based pruning is advantageous due to its faster ranking computation.





\subsection{Retraining Budget}

In this section, we consider the retraining budget alongside pruning rate and accuracy. We pose the question: \textit{For a given pruning rate and computational budget, which method yields the best performance?} In Figure~\ref{fig:retraining_budget} we present three plots representing different pruning rates, comparing the budgets used by one-shot and iterative pruning to achieve a given accuracy. The results shown here are based on the ResNet architecture trained on CIFAR-10; additional examples can be found in the Appendix.

The retraining budget is measured in terms of the total number of retraining epochs. For one-shot pruning, this budget corresponds to a single sequence of epochs. For iterative pruning, it represents the sum of epochs over all iterations. As illustrated in Figure~\ref{fig:retraining_budget}, one-shot pruning proves to be the most efficient approach for pruning rates up to 80\% across all computational budgets, achieving higher accuracy across the range of total epochs. However, at higher pruning rates, iterative pruning shows improved performance, making it the preferred method in these cases.



\subsection{Hybrid Regime}
\label{sec:hybrid}

The findings from this work indicate that for lower pruning ratios, one-shot pruning is generally more effective than iterative pruning. Building on this insight, we propose a hybrid few-shot approach that combines elements of one-shot and iterative pruning. This hybrid method prunes a large portion of the network in a one-shot-like step, followed by a more refined, geometric pruning strategy.
The results, shown in Figure~\ref{fig:hybrid}, demonstrate that the hybrid approach performs best across nearly all pruning rates, particularly enhancing performance at lower pruning rates. Hybrid pruning leverages the strengths of both one-shot and iterative approaches: it removes the majority of weights in the initial iteration, reducing redundant cycles early in the pruning process, while retaining the precision of geometric iterations at higher pruning rates, where remaining weights carry greater importance and require finer adjustment.

Benchmarking the hybrid approach provides valuable insights into optimal parameter settings. As a general guideline, in the initial step, 60–80\% of the target pruning rate \( p \) can be pruned (denoted as \( p_k \)), followed by retraining with extended patience (approximately 200 epochs). The remaining weights are then pruned iteratively with a rate \( p_i \ll p_k \), using diminishing amounts defined by a geometric sequence. For final pruning rates $p <80\%$, the iterative pruning rate \( p_k \) can be around 10\%, while for higher pruning rates \( p_k \) decreases to about 2\%. Fine-tuning then continues with patience set to approximately $\frac{1}{20}$ of the patience used in one-shot pruning. 

\section{Conclusion}

In summary, this study provides a broad evaluation of one-shot and iterative pruning strategies, addressing a critical gap in neural network optimization research. While one-shot pruning is effective at lower pruning ratios, iterative pruning proves superior for higher pruning rates, and arguably transformer architectures and second-derivative pruning criteria. Additionally, our proposed hybrid pruning integrates the strengths of both one-shot and iterative approaches.

This study offers an empirical basis for practitioners to select a pruning regime, including key hyperparameters such as pruning length, incorporating a proposed patience-based approach and step size. Choosing an optimal pruning strategy should be tailored to the specific performance objectives and computational constraints. Future research should further investigate the impact of pruning strategies under different pruning criteria, addressing limitations identified in this work and refining techniques for more effective pruning regimes.

\section{Acknowledgments}
We gratefully acknowledge Polish high-performance computing infrastructure PLGrid (HPC Center: ACK Cyfronet AGH) for providing computer facilities and support within computational grant no. PLG/2024/017173. The work of Tomasz Wojnar was supported by the National Centre of Science (Poland) Grant No. 2023/50/E/ST6/00068. The work of Mikołaj Janusz was funded by the "Interpretable and Interactive Multimodal Retrieval in Drug Discovery" project. The „Interpretable and Interactive Multimodal Retrieval in Drug Discovery” project (FENG.02.02-IP.05-0040/23) is carried out within the First Team programme of the Foundation for Polish Science co-financed by the European Union under the European Funds for Smart Economy 2021-2027 (FENG).



\bibliography{arxiv}

\clearpage
\renewcommand{\thesection}{\Alph{section}} 
\section*{APPENDIX} 

\setcounter{section}{0}

\begin{figure*}[h]
\centering

\begin{subfigure}[b]{0.19\textwidth}
    \centering
    \includegraphics[width=\textwidth, trim=0 0 70 65, clip]{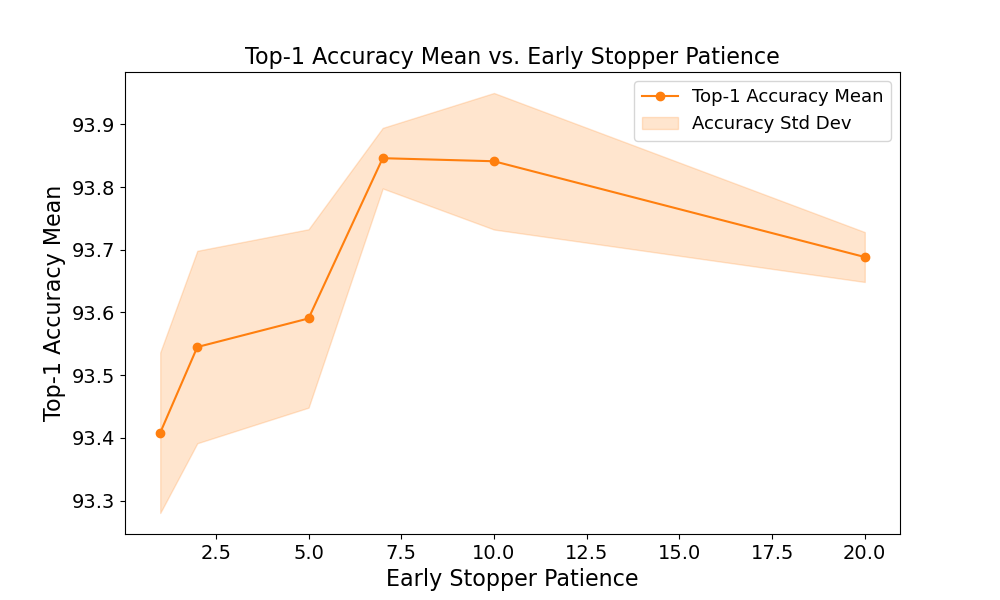}
    \caption{Iterative geometric pruning (varying patience)}
\end{subfigure}
\hfill
\begin{subfigure}[b]{0.19\textwidth}
    \centering
    \includegraphics[width=\textwidth, trim=0 0 70 65, clip]{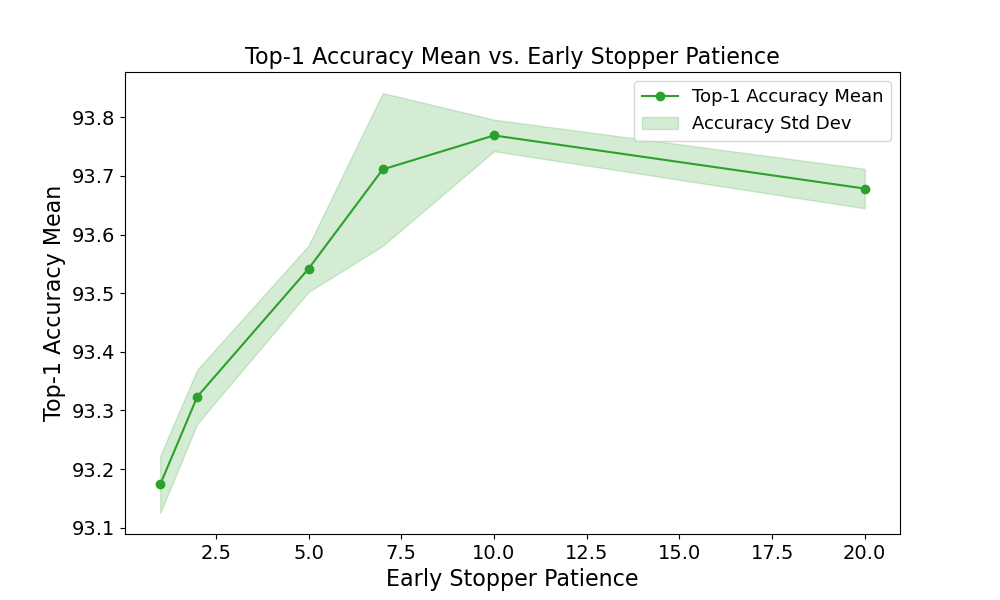}
    \caption{Iterative constant pruning (varying patience)}
\end{subfigure}
\hfill
\begin{subfigure}[b]{0.19\textwidth}
    \centering
    \includegraphics[width=\textwidth, trim=0 0 70 65, clip]{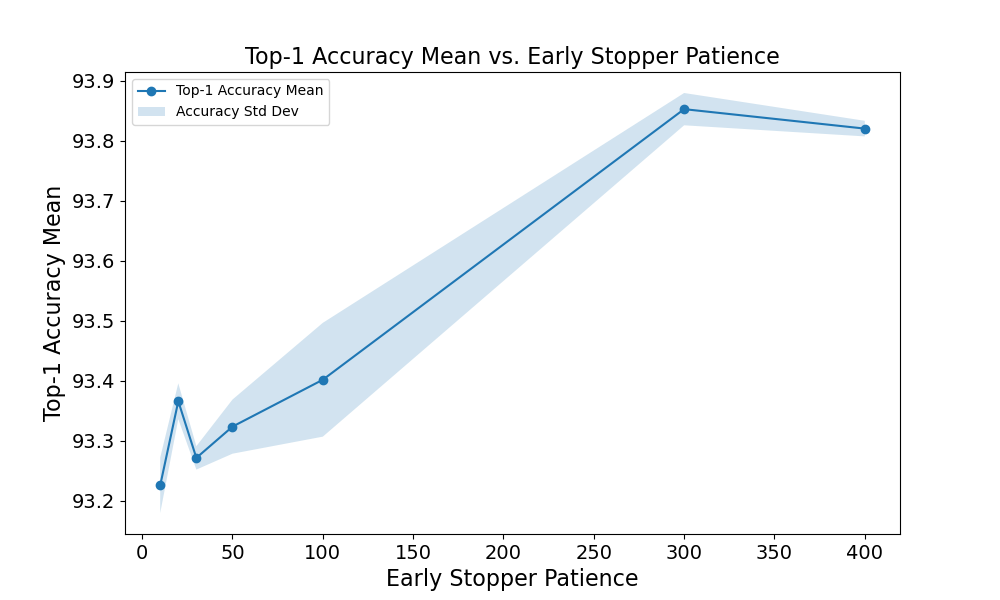}
    \caption{One-shot pruning (varying patience)}
\end{subfigure}
\hfill
\begin{subfigure}[b]{0.19\textwidth}
    \centering
    \includegraphics[width=\textwidth, trim=0 0 70 65, clip]{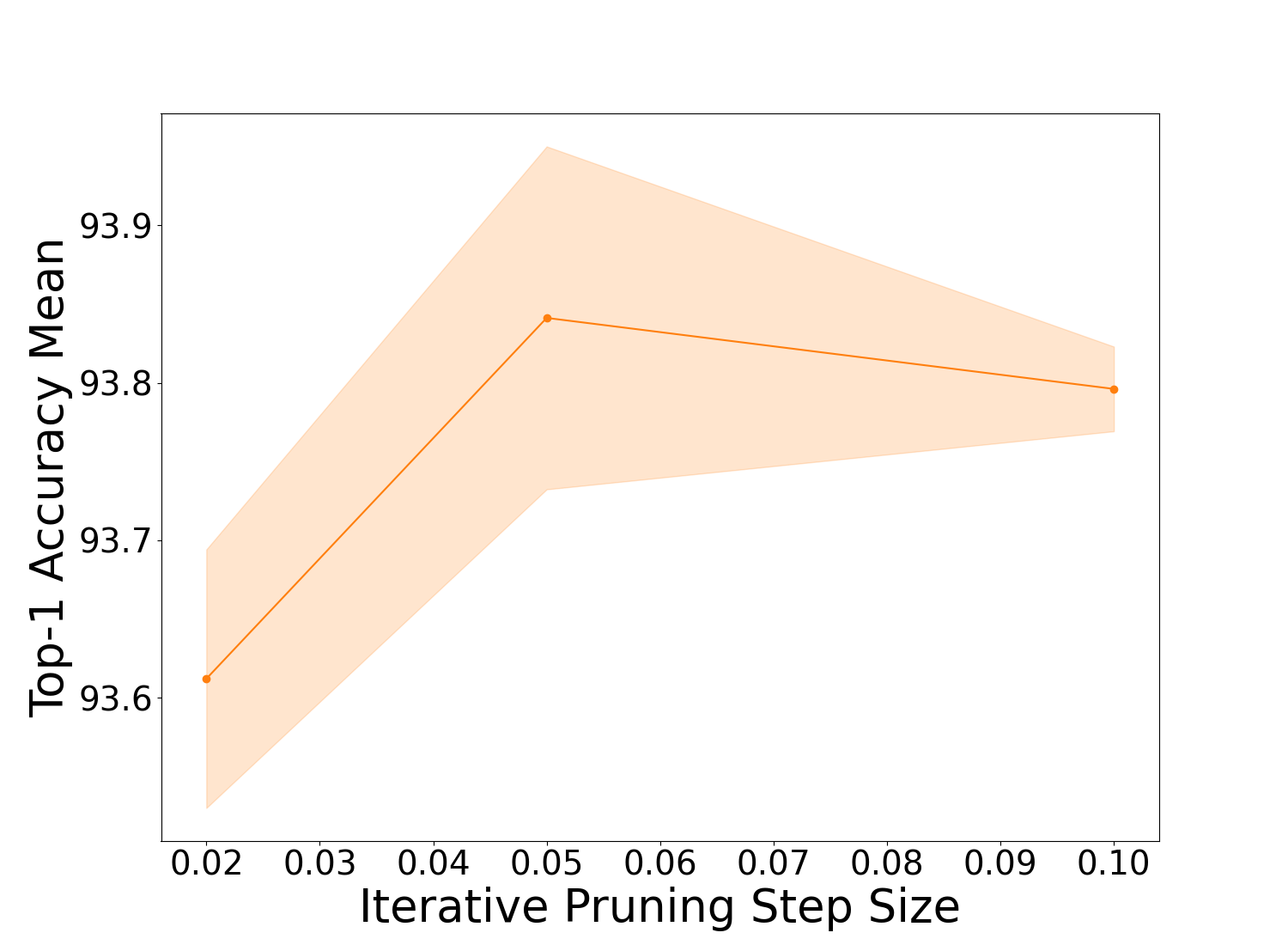}
    \caption{Iterative constant pruning (varying step size)}
\end{subfigure}
\hfill
\begin{subfigure}[b]{0.19\textwidth}
    \centering
    \includegraphics[width=\textwidth, trim=0 0 70 65, clip]{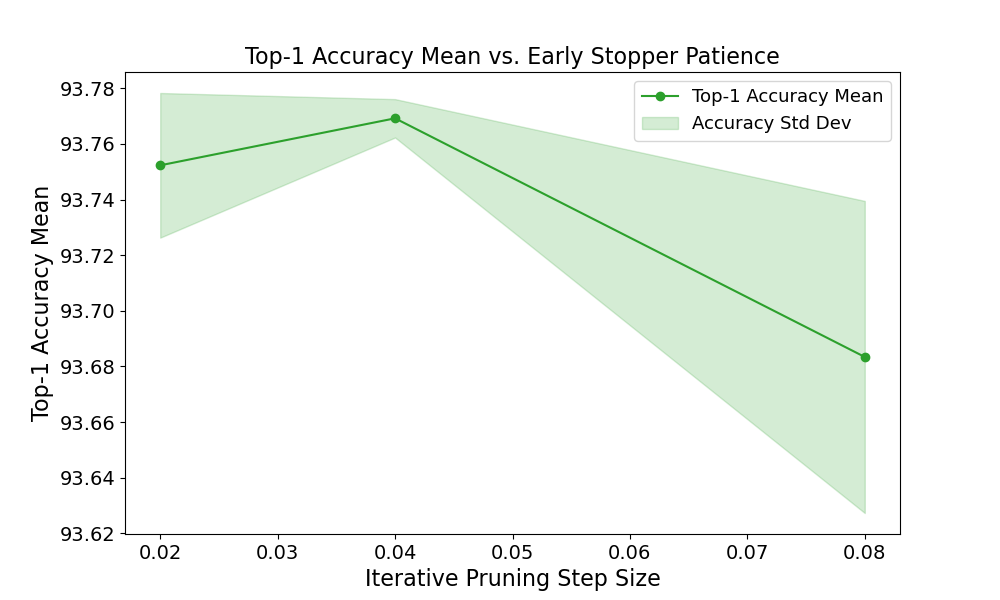}
    \caption{Iterative geometric pruning (varying step-size)}
\end{subfigure}
    \vspace{0.8cm}
\caption{Varying patience and step size (x-axis) impacts the pruning performance (y-axis).  In Fig. (a-c) patience is varied and in Fig. (d-e) step size is varied (only for iterative pruning regimes). All experiments are done for pruning rate 88\% and CIFAR-10 / ResNet-18.}
\label{fig:ablations}
    \vspace{0.8cm}
\end{figure*}

\section{Ablating pruning retraining length and step size.}
\label{sec:ablation}

As discussed in this work, the retraining duration is a crucial parameter for network pruning, influenced by two main factors: patience and step size. This ablation study highlights the substantial impact of these parameters on pruning performance. Patience-based training allows for better adaptation during retraining cycles. However, as shown in Fig.~\ref{fig:ablations}(a-c), choosing an arbitrary patience value can negatively affect performance. Extended retraining may not only waste computational resources but also lead to performance degradation. Similarly, step size is essential in iterative pruning, as illustrated in Fig.~\ref{fig:ablations}(d-e). Step size determines the frequency of pruning and fine-tuning cycles, and our findings reveal a non-monotonic trend: both overly small and excessively large steps can reduce performance.

\section{Pruning criteria further insights}

In this work, we experiment with several pruning criteria, including magnitude pruning (Fig.~\ref{fig:main_results}), Hessian-based pruning~\cite{lecun1990optimal} (Fig.~\ref{fig:tiny_crit}b-c; see Appendix for structured pruning adaptation), and Taylor expansion-based contribution approximation pruning~\cite{molchanov2019importance} (see Appendix).

Our primary focus is on magnitude pruning due to its simplicity, effectiveness, reliability, and low computational cost, allowing for extensive benchmarking and experimentation~\cite{hoefler2021sparsity}. Magnitude pruning is one of the most widely used pruning criteria across a variety of methods~\cite{han2015deep, han2015learning, renda2020comparing}. 

In this section, however, we want to address the potential similarities and differences in evaluating pruning regimes when alternative criteria are applied. Overall, we find that the relative performance of pruning regimes is largely consistent across different criteria. With the appropriate retraining duration, one-shot pruning performs best up to 80\% of the original parameter count, while iterative pruning is preferable at higher compression ratios. Notably, for Hessian-based criteria, one-shot pruning at high pruning rates results in a significant accuracy drop, suggesting iterative pruning may be a more stable solution for second-derivative-based methods.

For second-derivative pruning, the Hessian matrix, which captures the curvature of the loss function, identifies weights in low-curvature regions (small eigenvalues) as good pruning candidates. The experimental results may be explained by the fact that single-step pruning can dramatically alter the loss landscape, rendering the pre-pruning Hessian less accurate in assessing remaining weights. In contrast, iterative pruning enables recalculating the Hessian at each step, ensuring a more precise sensitivity evaluation of the weights retained.

\begin{figure*}[h!]
    \centering
    \begin{subfigure}[b]{0.48\linewidth}
        \centering
        \includegraphics[trim=0 0 0 50, clip, width=\linewidth]{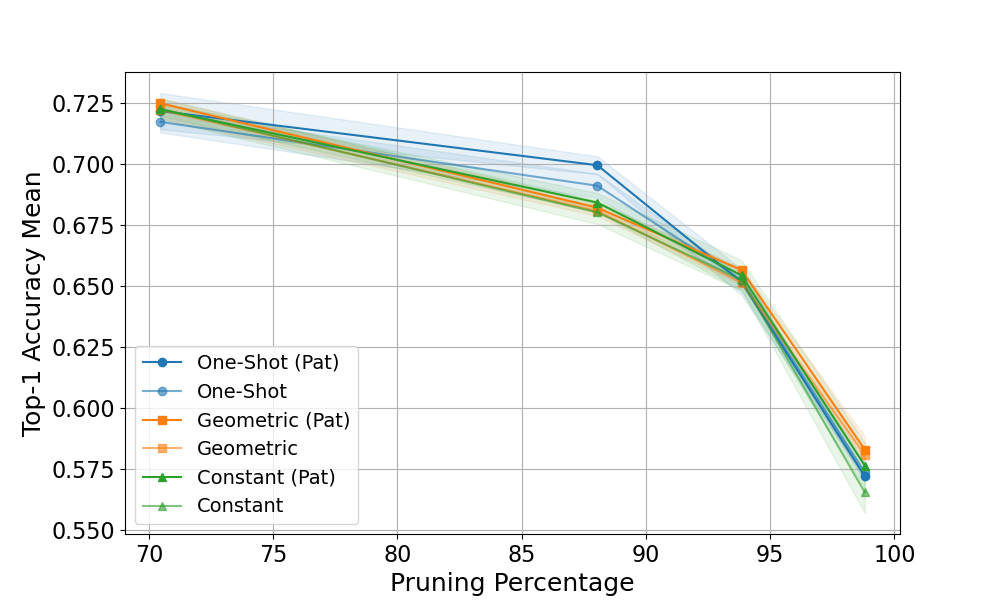}
        \caption{Hessian pruning (structured)}
        \label{fig:hessian_cif100_str}
    \end{subfigure}
    \hfill
    \begin{subfigure}[b]{0.48\linewidth}
        \centering
        \includegraphics[trim=0 0 0 50, clip, width=\linewidth]{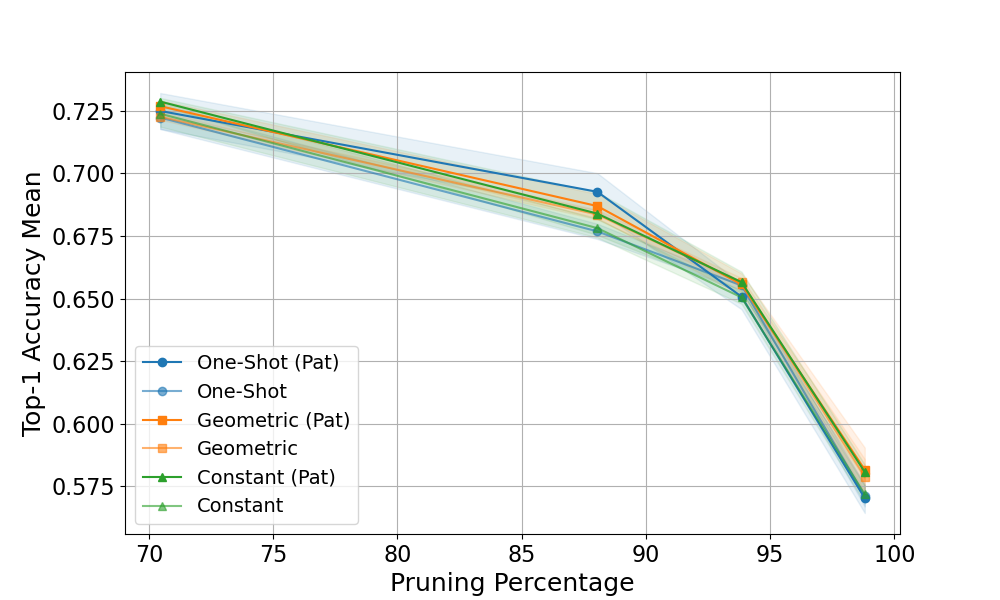}
        \caption{Taylor pruning (structured)}
        \label{fig:taylor_cif10_str}
    \end{subfigure}
        \vspace{0.8cm}
    \caption{Second-derivative (Hessian) pruning criteria. Iterative vs. one-shot pruning for CIFAR-100 and CIFAR-10.}
    \label{fig:othercritpruning_str}
    \vspace{0.8cm}
\end{figure*}

We then expand on the main text, which compares different pruning criteria under various training regimes. We present additional results on structured pruning using two criteria: Hessian-based pruning \cite{lecun1990optimal} and Taylor expansion-based contribution approximation pruning \cite{molchanov2019importance}. The results, shown in Fig.~\ref{fig:othercritpruning_str}, are largely consistent with the conclusions drawn in the main paper. Specifically, one-shot pruning performs better or comparably up to about a 90\% pruning rate, whereas iterative pruning yields better performance at higher compression ratios.


\subsection{Further magnitude-pruning results.}

Figure~\ref{fig:main_results} presents further comparison of one-shot and iterative pruning across various network architectures and vision datasets. The iterative pruning comes in two types: constant and geometric.

\begin{figure*}[tp]
\centering
\begin{subfigure}[b]{0.32\textwidth}
    \centering
    \includegraphics[width=\textwidth, trim=15 3 80 65, clip]{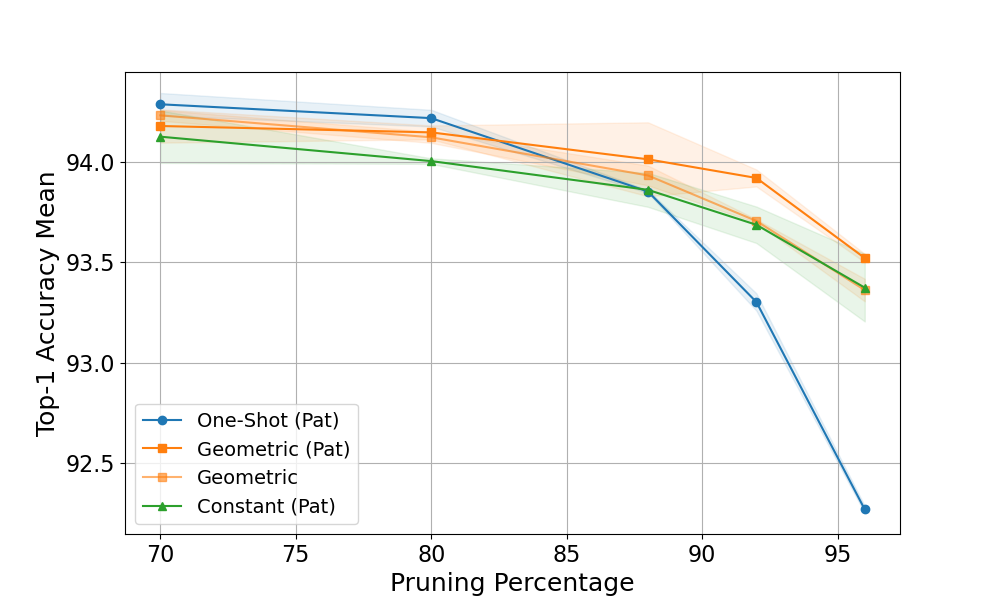}
    \caption{\makecell{ResNet-18 / CIFAR-10 \\ }}
\end{subfigure}
\hfill
\begin{subfigure}[b]{0.32\textwidth}
    \centering
    \includegraphics[width=\textwidth, trim=15 3 80 65, clip]{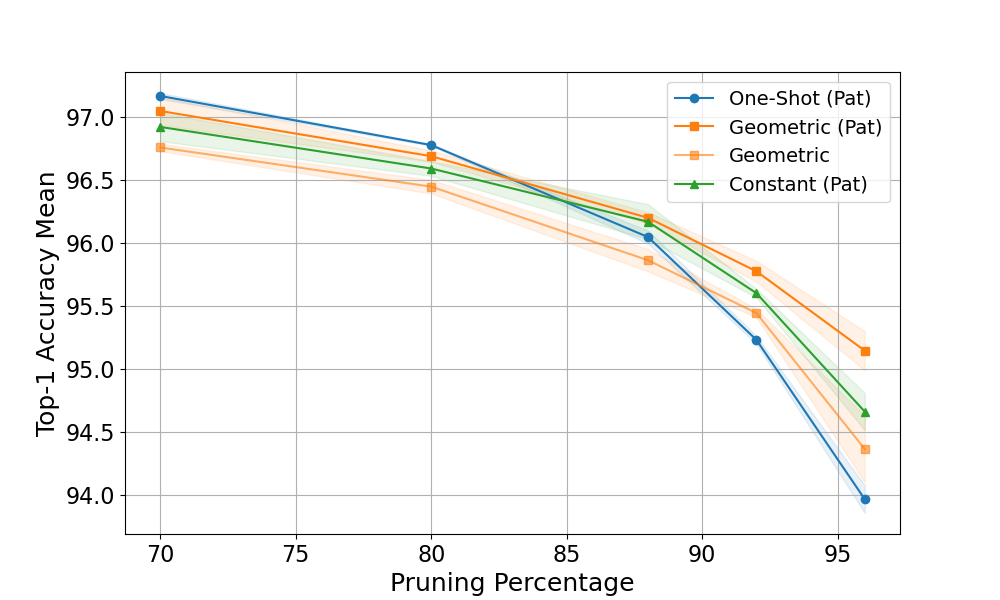}
    \caption{\makecell{EfficientNet / CIFAR-10 \\ }}
\end{subfigure}
\hfill
\vspace{1.5cm}
\begin{subfigure}[b]{0.32\textwidth}
    \centering
    \includegraphics[width=\textwidth, trim=15 3 80 65, clip]{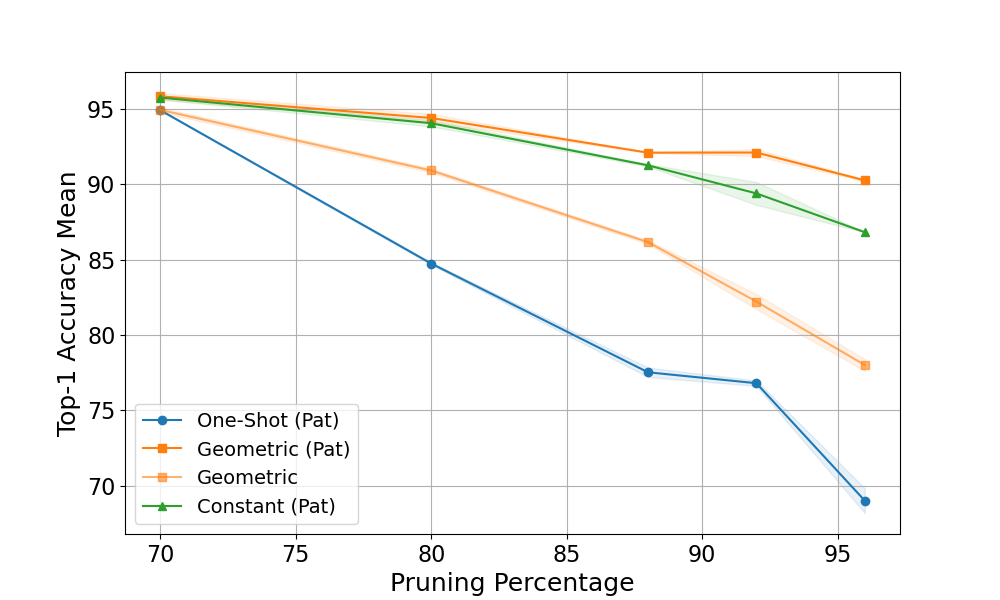}
    \caption{\makecell{ViT / CIFAR-10 \\ }}
\end{subfigure}
    \vspace{0.8cm}
\begin{subfigure}[b]{0.32\textwidth}
    \centering
    \includegraphics[width=\textwidth, trim=15 3 80 65, clip]{figures/plots1411/resnet18_cifar100_unstructured_all.png}
    \caption{\makecell{ResNet-18 / CIFAR-100 \\ }}
\end{subfigure}
\hfill
\begin{subfigure}[b]{0.32\textwidth}
    \centering
    \includegraphics[width=\textwidth, trim=15 3 80 65, clip]{figures/plots1411/efficientnetv2s_cifar100_unstructured_all.png}
    \caption{\makecell{EfficientNet / CIFAR-100 \\ }}
\end{subfigure}
\hfill
    \vspace{0.8cm}
\begin{subfigure}[b]{0.32\textwidth}
    \centering
    \includegraphics[width=\textwidth, trim=15 3 80 65, clip]{figures/plots1411/vit_cifar100_unstructured_all.png}
    \caption{\makecell{ViT / CIFAR-100 \\ }}
\end{subfigure}


\begin{subfigure}[b]{0.32\textwidth}
    \centering
    \includegraphics[width=\textwidth, trim=15 3 80 65, clip]{figures/plots1411/resnet18_imagenet1k_unstructured_all.png}
    \caption{\makecell{ResNet-18 / Imagenet \\ }}
\end{subfigure}
\hfill
\begin{subfigure}[b]{0.32\textwidth}
    \centering
    \includegraphics[width=\textwidth, trim=15 3 80 65, clip]{figures/with_patience/cifar10_structured_three_lr0.001_earlystopper_True_iterstepitNone_iterstepconstNone_iterftNone_norm2.png}
    \caption{\makecell{ResNet-18 / CIFAR-10 \\ (Structured)}}
\end{subfigure}
\hfill
\begin{subfigure}[b]{0.32\textwidth}
    \centering
    \includegraphics[width=\textwidth, trim=15 3 80 60, clip]{figures/plots1411/resnet18str_cifar100_structured_all.png}
    \caption{\makecell{ResNet-18 / CIFAR-100 \\ (Structured)}}
\end{subfigure}
\vspace{0.5cm}
\caption{Comparison of one-shot and iterative pruning across various network architectures and vision datasets.}
\vspace{0.5cm}
\end{figure*}

\section{Early stopping.}

In this section, we provide the detailed algorithm for early stopping performed in this paper. We use this algorithm for both one-shot and iterative geometric and iterative constant pruning. 

    \begin{algorithm}
\textbf{Note:} This code assumes that a lower metric value indicates better performance (e.g., loss). Otherwise, if a higher metric value is better (e.g., accuracy) the code is run with a reversed comparison.
\caption{Early Stopping Check}
\begin{algorithmic}[1]
\Procedure{EarlyStop}{$metric\_value: float$}
    \If{$metric\_value < self.best\_metric\_value$}
        \State $self.best\_metric\_value \gets metric\_value$
        \State $self.counter \gets 0$
    \ElsIf{$metric\_value > (self.best\_metric\_value + self.min\_delta)$}
        \State $self.counter \gets self.counter + 1$
    \EndIf
    \If{$self.counter \geq self.patience$}
        \State \Return True
    \EndIf
    \State \Return False
\EndProcedure
\end{algorithmic}
\label{algo_patience}
\end{algorithm}

\section{Structured and unstructured pruning}

\paragraph{Unstructured pruning.} Unstructured pruning involves selectively removing individual weights from the neural network based on criteria such as weight magnitude or their impact on the loss function~\cite{han2015deep,frankle2018lottery}. This method creates sparse weight matrices with many zero elements, which can significantly reduce the parameter count. However, practical computational gains often require specialized hardware or software optimizations because the remaining weights are irregularly distributed across the network.

\paragraph{Structured pruning.} In contrast, structured pruning removes entire components within the neural network, such as filters, channels, neurons, or even layers~\cite{he2018amc,ye2018rethinking,liu2019rethinking,li2020group,li2020dhp}. This method produces a more compact and regular network structure that retains a dense matrix format, making it easier to optimize on standard hardware. Structured pruning can substantially reduce both model size and computational requirements while maintaining a more organized network. However, achieving high sparsity ratios with structured pruning is more challenging, as it requires removing entire rows or columns rather than individual elements within a weight matrix.

The pruning regimes discussed in the following section apply to both unstructured and structured pruning. However, implementation details may vary due to constraints imposed by the structure of the pruned components. 

\subsection{Structured pruning pruning ratios}

We then provide details on the pruning percentages for each layer in structured pruning. In unstructured pruning, we perform global pruning, allowing pruning to occur freely in any layer. However, applying this approach to structured pruning can lead to pruning collapse, where a layer ends up without any channels. To prevent this, we define a separate pruning ratio for each layer. These ratios are chosen so that the total number of pruned channels across the entire network matches the desired overall pruning percentage. The details of each layer's pruning ratio and the final pruning percentages are given in Table \ref{tab:pruning_ratios}.

\begin{table*}[h!]
\centering
\begin{tabular}{lccccccl}
\toprule
\textbf{Layer Name} & \textbf{Conv1 (\%)} & \textbf{Layer1 (\%)} & \textbf{Layer2 (\%)} & \textbf{Layer3 (\%)} & \textbf{Layer4 (\%)} & \textbf{Pruning Ratio (\%)} \\ 
\midrule
Model 1             & 20                  & 20                  & 30                  & 40                  & 50                  & 69.61              \\ 
Model 2             & 50                  & 50                  & 60                  & 70                  & 80                  & 93.27               \\ 
Model 3             & 40                  & 40                  & 50                  & 60                  & 70                  & 85.25                     \\ 
Model 4             & 65                  & 65                  & 75                  & 85                  & 95                  & 97.63                     \\ 
\bottomrule
\end{tabular}
    \vspace{0.8cm}
\caption{Pruning percentages for layers and corresponding pruning ratios.}
\label{tab:pruning_ratios}
    \vspace{0.8cm}
\end{table*}

\section{Hybrid regime experimental details}

In the hybrid regime experiments, we used the same configuration as in the ResNet-18 experiments on CIFAR-100 and CIFAR-10 datasets. The hybrid regime consisted of an initial one-shot pruning step to a value of $p\%$, followed by iterative geometric steps with a ratio of $p_i$ until reaching the desired total pruning percentage. The first iterative step begins at $p_k\%$. For the hybrid regime, we used different patience values for the one-shot part and the iterative geometric part. We tested all the configurations provided in Tables \ref{tab:hybrid_scheduler_table_one} and \ref{tab:hybrid_scheduler_table_second}. For the sake of preciseness we provide the exact pruning percentages $p_k$ which were used in the iterative phase of the hybrid regime; however in this phase, we aimed to test a set of iterative percentages from the set, $p_k=0.02, 0.05, 0.10$. The adjustments were necessary to obtain the exact final pruning ration $p$ and the fair comparison with other pruning methods.

\begin{table}[h]
\centering
\caption{Pruning Hybrid Scheduler Parameters}
\label{tab:hybrid_scheduler_table_one}
\begin{tabular}{cccc}
\toprule
\textbf{One-shot step} & \textbf{Iterative step} & \textbf{Target pruning value} \\
\midrule
0.5 & 0.01842 & 0.7 \\
0.6 & 0.01741 & 0.7 \\
0.5 & 0.04365 & 0.7 \\
0.6 & 0.03451 & 0.7 \\
0.5 & 0.07168 & 0.7 \\
0.6 & 0.1 & 0.7 \\
0.5 & 0.01962 & 0.8 \\
0.6 & 0.01842 & 0.8 \\
0.7 & 0.01741 & 0.8 \\
0.5 & 0.04968 & 0.8 \\
0.6 & 0.04365 & 0.8 \\
0.7 & 0.03451 & 0.8 \\
0.5 & 0.08531 & 0.8 \\
0.6 & 0.07168 & 0.8 \\
0.7 & 0.05132 & 0.8 \\
0.5 & 0.01972 & 0.88 \\
0.6 & 0.01914 & 0.88 \\
0.7 & 0.01965 & 0.88 \\
0.8 & 0.01654 & 0.88 \\
0.5 & 0.04668 & 0.88 \\
0.6 & 0.04585 & 0.88 \\
0.7 & 0.0484 & 0.88 \\
0.8 & 0.04083 & 0.88 \\
0.5 & 0.09118 & 0.88 \\
0.6 & 0.07884 & 0.88 \\
0.7 & 0.09446 & 0.88 \\
0.8 & 0.08 & 0.88 \\
\bottomrule
\end{tabular}
\end{table}

\begin{table}[h]
\centering
\caption{Pruning Hybrid Scheduler Parameters 0.92 - 0.99}
\label{tab:hybrid_scheduler_table_second}
\begin{tabular}{cccc}
\toprule
\textbf{One-shot step} & \textbf{Iterative step} & \textbf{Target pruning value} \\
\midrule
0.5 & 0.01997 & 0.92 \\
0.6 & 0.0191 & 0.92 \\
0.7 & 0.01893 & 0.92 \\
0.8 & 0.0181 & 0.92 \\
0.5 & 0.04831 & 0.92 \\
0.6 & 0.04706 & 0.92 \\
0.7 & 0.04848 & 0.92 \\
0.8 & 0.04172 & 0.92 \\
0.5 & 0.08679 & 0.92 \\
0.6 & 0.09191 & 0.92 \\
0.7 & 0.07948 & 0.92 \\
0.8 & 0.06192 & 0.92 \\
0.5 & 0.01968 & 0.96 \\
0.6 & 0.01922 & 0.96 \\
0.7 & 0.01987 & 0.96 \\
0.8 & 0.01919 & 0.96 \\
0.5 & 0.04629 & 0.96 \\
0.6 & 0.04838 & 0.96 \\
0.7 & 0.04895 & 0.96 \\
0.8 & 0.04265 & 0.96 \\
0.5 & 0.0976 & 0.96 \\
0.6 & 0.08539 & 0.96 \\
0.7 & 0.0955 & 0.96 \\
0.8 & 0.08348 & 0.96 \\
0.5 & 0.01961 & 0.99 \\
0.6 & 0.01958 & 0.99 \\
0.7 & 0.01994 & 0.99 \\
0.8 & 0.01897 & 0.99 \\
0.5 & 0.04696 & 0.99 \\
0.6 & 0.04823 & 0.99 \\
0.7 & 0.04775 & 0.99 \\
0.8 & 0.04127 & 0.99 \\
0.5 & 0.09171 & 0.99 \\
0.6 & 0.09413 & 0.99 \\
0.7 & 0.08206 & 0.99 \\
0.8 & 0.1 & 0.99 \\
\bottomrule
\end{tabular}
\end{table}

\section{Experiments set-up}
\subsection{Dependencies}
The technical setup for the experiments included the following dependencies:
\begin{itemize}
    \item Python 3.11
    \item CUDA 12.1
    \item PyTorch 2.2
    \item Torchvision 0.17.0
    \item timm 0.9.16
    \item torch-pruning 1.4.2
\end{itemize}
Experiments were conducted mainly on NVIDIA A100 and RTX 2080ti GPU's.

\subsection{Dataset transformations}
Cifar-10 and Cifar-100 transformations include normalization (values are located in the code repository), random crop of size 32×32 with padding 4 and random horizontal flip. The images were also resized to higher resolution for some models. ImageNet1K was normalized, resized and cropped.




\subsection{Checkpoints}
The checkpoints for the models used in
the experiments are shared here: \url{https://www.dropbox.com/scl/fo/u0d8a087o3c2ynzpb6chd/AJz5w2ozXzcrBzxUwXVMiYM?rlkey=gag0w2r89kmt1huek6zsy9re2&st=4guxofag&dl=0}.

\subsection{Parameters}
All experiments were conducted using the SGD optimizer with the following parameters:
\begin{itemize}
    \item Learning rate: 0.01
    \item Momentum: 0.9
    \item Weight decay: 0.0005
\end{itemize}

The batch size was set to 512, consistent across all pruning experiments. The training data was shuffled for every run.

\subsubsection{CIFAR-100 / ResNet18}
We used the ResNet-18 model from the following GitHub repository:  
\url{https://github.com/kuangliu/pytorch-cifar/blob/master/models/resnet.py}

Before pruning, the model was trained for 328 epochs with the following configuration:
\begin{itemize}
    \item SGD optimizer with learning rate: 0.1, momentum: 0.9, and weight decay: 0.0005
    \item Linear scheduler for 100 epochs with start factor: 0.01
    \item After 100 epochs: CosineAnnealingWarmRestarts with $T_0 = 50$, $T_{\text{mult}} = 2$, and $\eta_{\text{min}} = 1 \times 10^{-5}$
    \item Early stopping with patience of 100 epochs
\end{itemize}
Before pruning, the top-1 accuracy was 74.64\%.

\subsubsection{CIFAR-10 / ResNet18}
The same ResNet-18 model as for CIFAR-100 was used.  
Before pruning, the model was trained for 226 epochs with the same configuration as the CIFAR-100 experiment.  
Before pruning, the top-1 accuracy was 94.14\%.

\subsubsection{ImageNet / ResNet18}
We used the ResNet-18 model from the PyTorch torchvision library with pretrained weights. During fine-tuning in the pruning phase, images were resized to $256 \times 256$ and cropped to $224 \times 224$.  
Before pruning, the top-1 accuracy was 68.91\%.

\subsubsection{CIFAR-100 / EfficientNet}
We used the EfficientNet V2-S model from the PyTorch torchvision library with pretrained weights (IMAGENET1K\_V1).  
Before pruning, the model was trained for 132 epochs with the following configuration:
\begin{itemize}
    \item SGD optimizer with learning rate: 0.1, momentum: 0.9, and weight decay: 0.0005
    \item Linear scheduler for 10 epochs with start factor: 0.01
    \item After 10 epochs: CosineAnnealingWarmRestarts with $T_0 = 10$, $T_{\text{mult}} = 2$, and $\eta_{\text{min}} = 1 \times 10^{-5}$
    \item Early stopping with patience of 80 epochs
    \item Images resized to $128 \times 128$
\end{itemize}
Images were resized during fine-tuning in the pruning phase. Before pruning, the top-1 accuracy was 87.53\%.

\subsubsection{CIFAR-10 / EfficientNet}
The same EfficientNet V2-S model as for CIFAR-100 was used.  
Before pruning, the model was trained for 152 epochs with the same configuration as the CIFAR-100 experiment.  
Before pruning, the top-1 accuracy was 97.88\%.

\subsubsection{CIFAR-100 / ViT}
We used the ViT small patch16 224 model from the timm library with pretrained weights (vit small patch16 224 augreg in1k).  
Before pruning, the model was trained for 18 epochs with the following configuration:
\begin{itemize}
    \item SGD optimizer with learning rate: 0.1, momentum: 0.9, and weight decay: 0.0005
    \item Linear scheduler for 10 epochs with start factor: 0.01
    \item After 10 epochs: CosineAnnealingWarmRestarts with $T_0 = 10$, $T_{\text{mult}} = 2$, and $\eta_{\text{min}} = 1 \times 10^{-5}$
    \item Early stopping with patience of 50 epochs
    \item Images resized to $224 \times 224$
\end{itemize}
Images were resized during fine-tuning in the pruning phase. Before pruning, the top-1 accuracy was 88.16\%.

\subsubsection{CIFAR-10 / ViT}
The same ViT small patch16 224 model as for CIFAR-100 was used.  
Before pruning, the model was trained for 19 epochs with the same configuration as the CIFAR-100 experiment.  
Before pruning, the top-1 accuracy was 98.11\%.

\end{document}